\pdfoutput=1

\documentclass[10pt,twocolumn,letterpaper]{article}

\usepackage[pagenumbers]{cvpr} 

\ifdefined\leftlinenumbers
  \leftlinenumbers*
\fi

\definecolor{cvprblue}{rgb}{0.21,0.49,0.74}
\usepackage[pagebackref,breaklinks,colorlinks,allcolors=cvprblue]{hyperref}

\title{Towards Generalized Spatio-Temporal Grounding in Long Videos: The LV-MOSTG-9K Benchmark and the LOMOST Model}

\author{
Chenhe Wang$^{1}$\quad
Linhui Xiao$^{2,*}$\quad
Kewei Feng$^{1}$\quad
Kaiyuan Duan$^{1}$\\
Chaodong Han$^{3}$\quad
Guoliang Chen$^{1}$\quad
Fengquan Zhang$^{1,*}$\\[0.25em]
$^{1}$Beijing University of Posts and Telecommunications\\
$^{2}$Pengcheng Laboratory\\
$^{3}$Communication University of China\\[0.2em]
{\small \texttt{chenhewang@bupt.edu.cn}, \texttt{xiaolinhui16@mails.ucas.ac.cn}, \texttt{zhangfq@bupt.edu.cn}}\vspace{-16pt}
}

\begin{document}
\makeatletter
\patchcmd{\@maketitle}{\vskip .375in}{\vskip .18in}{}{}
\patchcmd{\@maketitle}{\vspace*{24pt}}{\vspace*{10pt}}{}{}
\patchcmd{\@maketitle}{\vspace*{12pt}}{\vspace*{3pt}}{}{}
\makeatother
\maketitle
\raggedbottom
{\renewcommand{\thefootnote}{\fnsymbol{footnote}}\footnotetext[1]{Corresponding authors.}}
\patchcmd{\endabstract}{\vspace*{12pt}}{\par}{}{}
\begin{abstract}
Given a natural language expression and a video, spatio-temporal video grounding (STVG) aims to identify the temporal segments and spatial locations of the described objects. Existing benchmarks mostly focus on short videos or assume matching targets exist, leaving zero-target and multi-target scenarios in long videos underexplored. In these scenarios, existing methods still struggle with modeling target counts, achieving accurate spatio-temporal localization, and the high cost of dense video frame encoding. To address these challenges, we introduce generalized spatio-temporal grounding in long videos and construct the LV-MOSTG-9K benchmark. To our knowledge, it is the first STVG benchmark to jointly evaluate zero, single, and multiple target expressions in continuous 5--10-minute videos. It comprises 8,788 expressions across 2,369 clips spanning films, documentaries, daily life, and travel, with 4.42 million bounding boxes annotated at the original frame rate. To bridge global long-video understanding with fine-grained spatial localization, we propose LOMOST, a coarse-to-fine grounding framework. LOMOST first predicts the target interval and coarse anchor-frame boxes from sparsely sampled frames spanning the full video. Within the predicted interval, its Spatial Localization Calibrator combines Semantic Query Contextual Modeling and Feature Prior Consistency Modulation, integrating query-conditioned contextual evidence with cross-frame feature consistency priors to refine frame-level target boxes. We evaluate LOMOST on three STVG benchmarks and achieve state-of-the-art spatio-temporal grounding performance on BOSTVG and LV-MOSTG-9K.
We will release the LOMOST code and the LV-MOSTG-9K dataset at \url{https://github.com/chenhe-hehe/LOMOST}.
\end{abstract}

\section{Introduction}
\label{sec:intro}

\begin{figure}[!t]
  \centering
  \includegraphics[width=\columnwidth]{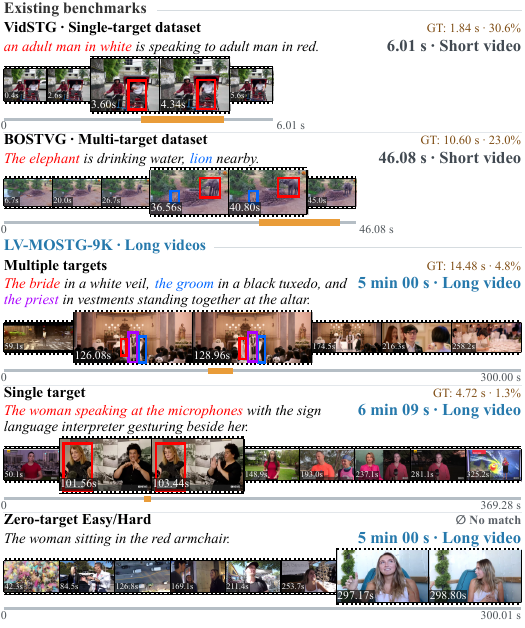}
  \caption{Representative queries from VidSTG, BOSTVG, and LV-MOSTG-9K. Orange bars mark ground-truth intervals on timelines scaled to each video's duration. Colored boxes show matching objects in selected frames. The two LV-MOSTG-9K queries with matches occupy 4.8\% and 1.3\% of their videos. Easy and hard zero-target queries both have no match.}
  \label{fig:teaser}
\end{figure}

Spatio-temporal video grounding (STVG) identifies when and where referred objects appear in a video~\cite{xiao2025towards}. Early benchmarks such as VidSTG and HC-STVG pair each expression with a single-object trajectory, jointly evaluating temporal boundaries and frame-level spatial localization~\cite{zhang2020does,tang2021human}. DVD-ST and BOSTVG extend this setting to multiple objects per expression~\cite{ji2024described,yao2026omnistvg}. When expressions describe actions and interactions, models must integrate cross-frame evidence to distinguish candidate instances and recover all referents.

In the image domain, generalized referring expression comprehension and segmentation extend grounding beyond a single referent to variable-sized target sets, including unmatched expressions~\cite{he2023grec,liu2023gres,hu2023beyond,hemanthage2024recantformer,wang2025hierarchical}.

Long-video understanding benchmarks such as MLVU and LongVideoBench assess whether models can find and use evidence across extended temporal contexts to answer questions~\cite{zhou2025mlvu,wu2024longvideobench}. In spatio-temporal grounding, the relevant segment may last only a few seconds, while visually similar objects and scenes can recur elsewhere in the video.

Existing long-video grounding benchmarks explore different query formulations: VUE-STG pairs each expression with a single target~\cite{josevidi2}, while Event-STVG uses an event type and occurrence index to localize specified roles~\cite{zhang2026event}. However, both assume matching targets and do not evaluate no-match recognition. Since queries in long videos may match zero, one, or multiple objects, evaluation should assess both no-match recognition and spatio-temporal localization of all matches.

Inspired by GREC and its extensions to video segmentation~\cite{he2023grec,liu2026sekai,ding2025mevis}, we introduce generalized spatio-temporal grounding in long videos (GL-STVG) and construct the Long-Video Multi-Object Spatio-Temporal Grounding benchmark (LV-MOSTG-9K). Given a 5--10-minute video and a referring expression, the task requires rejecting unmatched queries or localizing all matching targets in time and space. Fig.~\ref{fig:teaser} illustrates the zero-, single-, and multi-target settings with brief target intervals.

Existing methods also struggle in this setting. Single-target methods such as TubeDETR, STCAT, CG-STVG, and TA-STVG~\cite{yang2022tubedetr,jin2022embracing,gu2024context,gu2025knowing} require adaptation to recognize unmatched queries and localize variable-sized target sets, while OmniTube~\cite{yao2026omnistvg} supports multi-target grounding in short videos without zero-target recognition in its original setting. Moreover, brief events and similar distractors in long videos degrade spatio-temporal localization, whereas dense full-video encoding incurs high inference cost.

To bridge global long-video understanding with fine-grained spatial localization, we propose \textbf{LOMOST} (LOng-video Multi-Object Spatio-Temporal grounding), a coarse-to-fine grounding framework. LOMOST predicts the target interval and coarse anchor-frame boxes from sparsely sampled frames spanning the full video, restricting subsequent dense localization to the predicted interval. Using these predictions as temporal and spatial priors, its Spatial Localization Calibrator combines \textbf{Semantic Query Contextual Modeling (SQCM)} and \textbf{Feature Prior Consistency Modulation (FPCM)} for frame-level spatial refinement. Visual Memory-Guided Query Refinement integrates object semantics, coarse positional cues, and visual memory to form queries for SQCM. SQCM uses these refined queries to evaluate candidate locations through query-conditioned spatial context, while FPCM modulates candidate responses based on feature consistency with anchor-frame priors. Their scores are fused over a shared candidate grid, integrating contextual evidence with cross-frame feature consistency priors. A Box Refinement Head then uses the fused scores and local features to update target centers and sizes within the predicted interval. Our contributions are threefold.
\begin{itemize}
    \item We introduce GL-STVG and construct LV-MOSTG-9K, a benchmark that evaluates expressions with zero, one, or multiple matching objects in continuous 5--10-minute videos.
    \item We propose LOMOST, a coarse-to-fine model that uses sparse global sampling to predict the target interval and coarse anchor-frame boxes. Within that interval, its spatial localization calibrator fuses FPCM and SQCM scores, which guide the box refinement head.
    \item We evaluate LOMOST on VidSTG, BOSTVG, and LV-MOSTG-9K, achieving state-of-the-art spatio-temporal grounding performance on BOSTVG and reaching 30.27\% m-vIoU on LV-MOSTG-9K.
\end{itemize}

\flushbottom
\section{Related Work}
\label{sec:related-work}

\paragraph{Visual and Spatio-Temporal Grounding.}
Image grounding localizes language-referred regions~\cite{yu2018mattnet,deng2021transvg,zhu2022seqtr}, with language-conditioned refinement and modulation~\cite{deng2023transvg++,ye2022shifting,xiao2024hivg,yao2024visual} and grounded pre-training~\cite{li2022grounded,liu2024grounding} supporting visual-language alignment.
Spatio-temporal video grounding methods combine language with visual evidence across frames to predict both the target interval and spatial locations~\cite{su2021stvgbert,wasim2024videogrounding}. STGRN uses a spatio-temporal relation graph to model relationships between objects within frames and changes across frames~\cite{zhang2020does}. Among later Transformer methods, TubeDETR uses slow and fast branches to process frames at different sampling rates~\cite{yang2022tubedetr}, STCAT uses a global multimodal template to guide predictions across frames~\cite{jin2022embracing}, and CG-STVG uses instance context to capture appearance and motion cues~\cite{gu2024context}. Related work studies semantic-role-based video object grounding~\cite{sadhu2020video} and referring video object segmentation~\cite{seo2020urvos,wu2023onlinerefer}. For multi-object grounding, Ji et al.\ adapt TubeDETR and STCAT to predict multiple spatio-temporal tubes~\cite{ji2024described}, while OmniTube uses language-guided object queries to localize multiple referents in a single expression~\cite{yao2026omnistvg}. For long videos, ART-STVG processes frames in temporal order and uses spatial and temporal memory to retain past evidence for grounding~\cite{gu2026towards}.

\paragraph{Vision-Language Models for Grounding.}
Multimodal language models connect language with spatial regions~\cite{chen2023shikra,peng2024grounding,you2024ferret,li2024groundinggpt,ma2024groma} and pixel-level masks~\cite{lai2024lisa,rasheed2024glamm}.
Referring expressions that combine actions, spatial relations, and interactions require models to distinguish visually similar instances using language and video context. CurrSTVG builds a dataset of compositional expressions organized by difficulty and trains a grounding model with curriculum-based reinforcement learning~\cite{wang2026learning}. In the related task of referring multi-object tracking, ReferGPT uses 3D position and motion cues to generate object descriptions, then matches them to queries in a zero-shot setting~\cite{chamiti2025refergpt}. STVG-R1 uses objects with consistent identifiers across frames as visual prompts and localizes targets by selecting instances~\cite{zhang2026stvg}. Vidi2 supports video question answering and spatio-temporal grounding in a single model that directly predicts target intervals and bounding boxes~\cite{josevidi2}. Bridge-STG uses queries to pass semantic information from temporal reasoning in a multimodal language model to a dedicated spatial decoder~\cite{tu2026bridging}.

\paragraph{Efficient Long-Video Processing.}
Visual token compression reduces long-video processing costs. FastV and SparseVLM select tokens by attention or text relevance~\cite{chen2024image,zhang2024sparsevlm}; DivPrune promotes feature diversity~\cite{alvar2025divprune}; VisionZip retains salient tokens and aggregates others~\cite{yang2025visionzip}. FrameFusion and HoliTom merge similar tokens~\cite{fu2025framefusion,shao2505holitom}, while SToP accounts for attention sinks during pruning~\cite{kim2026sink}. SemVID allocates budgets using query relevance and frame changes, retaining object, motion, and context cues~\cite{li2026keeping}. We compare five strategies at three retention rates to assess their effects on fine-grained grounding (Sec.~\ref{sec:token-compression}).

\section{LV-MOSTG-9K Benchmark}
\label{sec:benchmark}

\subsection{Task Definition and Design Principles}
\label{sec:task-definition}

\paragraph{Task definition.}
In GL-STVG, the model receives a long video $V=\{(\boldsymbol{I}_i,t_i)\}_{i=1}^{N}$ and a natural-language referring expression $q$. Here, $\boldsymbol{I}_i$ and $t_i$ denote the image and timestamp of frame $i$, respectively, and $N$ is the number of video frames. The model determines whether any object in the full video matches the expression and returns an empty result if there is no match. Each expression with a match is associated with a single, semantically coherent target interval $\tau=[s,e)$. The model predicts the interval $\hat{\tau}=[\hat{s},\hat{e})$ and outputs a set of bounding boxes $\hat{\mathcal{B}}_i=\{\hat{\boldsymbol{b}}_{i,j}\}_{j=1}^{\hat{n}_i}$ for all visible matching objects on each evaluation frame satisfying $\hat{s}\leq t_i<\hat{e}$, where $\hat{n}_i$ is the predicted number of objects and $\hat{\boldsymbol{b}}_{i,j}$ is the $j$-th predicted box on frame $i$. The set may vary with target visibility and may be empty on individual frames. Cross-frame instance identities are optional.

\paragraph{Design principles.}
LV-MOSTG-9K follows six design principles.
\textit{(i) Scenario and category diversity:} Videos span films, documentaries, and vlogs, covering people, animals, plants, vehicles, and everyday objects.
\textit{(ii) Semantic richness:} Expressions describe attributes, actions, state changes, and interactions requiring contextual understanding.
\textit{(iii) Same-category instance selection:} Expressions distinguish subsets of same-category instances through spatial, appearance, action, or relational cues.
\textit{(iv) Generalized reference:} Following GREC~\cite{he2023grec}, queries refer to zero, one, or multiple matching objects.
\textit{(v) Long videos with dense annotations:} Continuous 5--10-minute videos retain background and distractors, with visible targets annotated at the original frame rate within target intervals.
\textit{(vi) Temporal uniqueness and coherence:} Each matched query identifies one semantically coherent interval; ambiguous queries are revised or excluded.

\paragraph{Comparison with existing benchmarks.}
As shown in Tab.~\ref{tab:benchmark-comparison}, BOSTVG evaluates queries with one or more targets, and VUE-STG pairs each query with a single target. Event-STVG specifies an event by type and occurrence order and asks for its designated roles. LV-MOSTG-9K uses referring expressions that may match zero, one, or multiple objects across a continuous 5--10-minute video.

\begin{table}[t]
  \centering
  \caption{Comparison of STVG benchmarks. Durations describe input videos or evaluation contexts.}
  \label{tab:benchmark-comparison}
  \small
  \setlength{\tabcolsep}{3pt}
  \begin{tabular}{@{}lll@{}}
    \toprule
    Benchmark & Input duration & Query targets \\
    \midrule
    VidSTG~\cite{zhang2020does} & $\sim$28 s (mean) & One object \\
    BOSTVG~\cite{yao2026omnistvg} & $\sim$37 s (mean) & One or more \\
    VUE-STG~\cite{josevidi2} & $\sim$10 s--30 min & One object \\
    Event-STVG~\cite{zhang2026event} & Up to 10 min & Specified event roles \\
    \textbf{LV-MOSTG-9K} & 5--10 min & Zero, one, or multiple \\
    \bottomrule
  \end{tabular}
\end{table}

\begin{figure*}[t]
  \centering
  \newsavebox{\datasetstatsleft}
  \sbox{\datasetstatsleft}{%
    \includegraphics[width=0.70\textwidth,trim=0 8.2bp 0 0,clip]{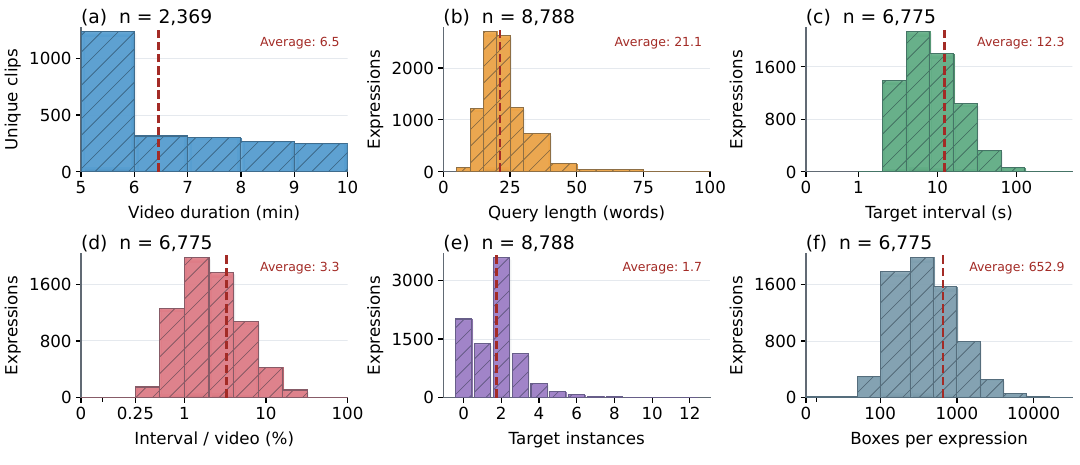}}
  \usebox{\datasetstatsleft}\hfill
  \resizebox{!}{\ht\datasetstatsleft}{%
    \vbox{%
      \hbox{\includegraphics[width=0.285\textwidth]{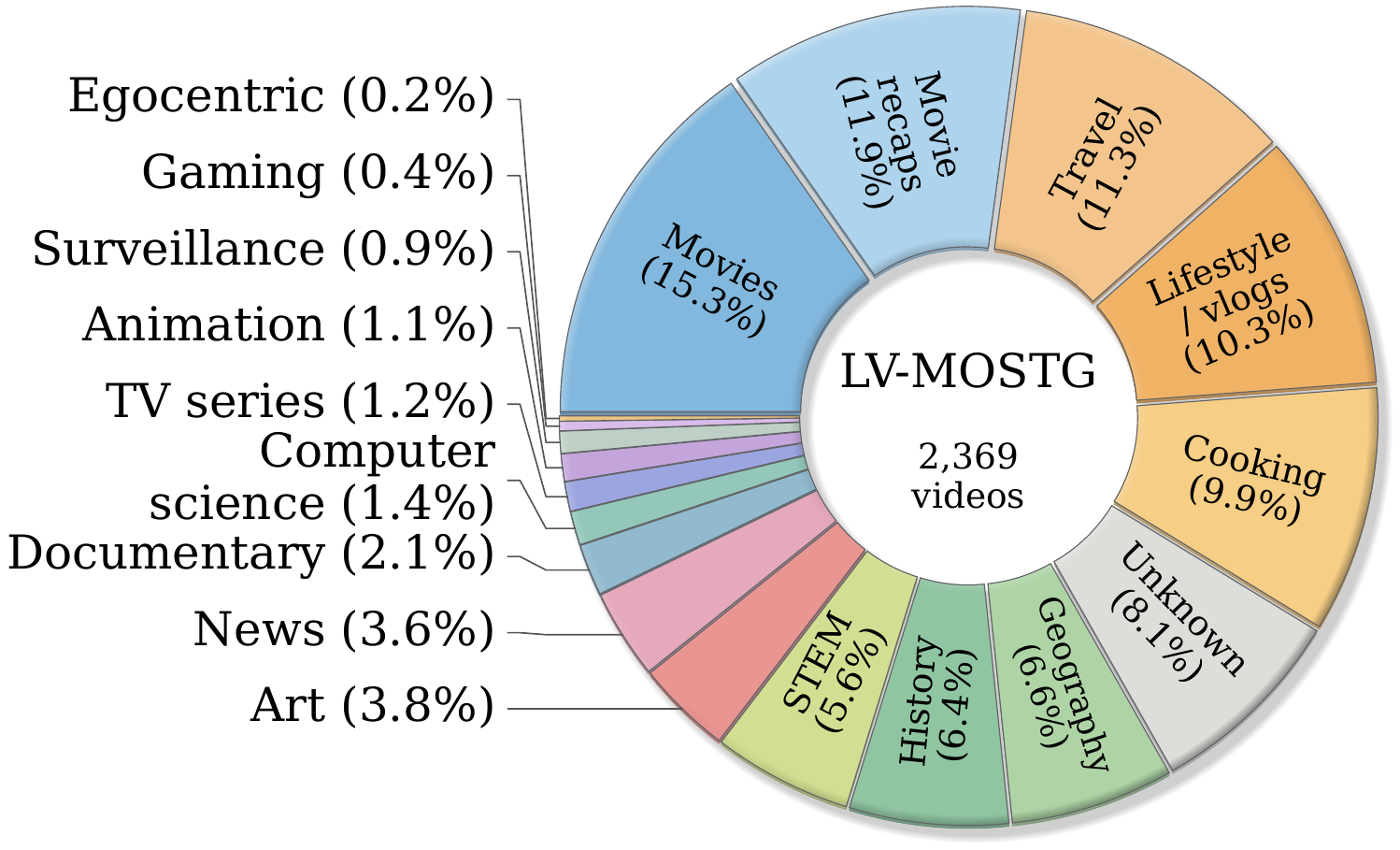}}%
      \kern 3pt
      \hbox{\includegraphics[width=0.285\textwidth]{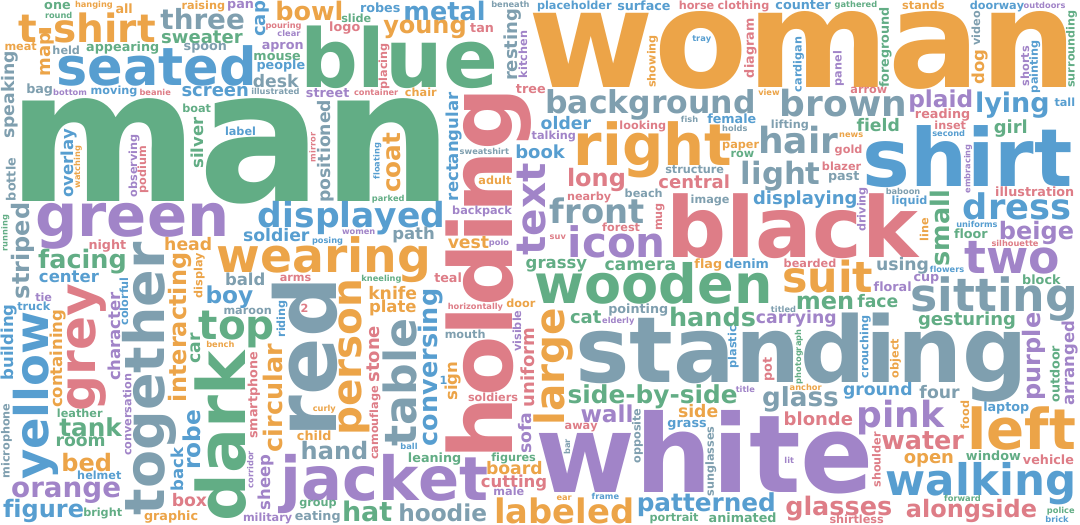}}%
    }}
  \caption{LV-MOSTG-9K statistics. Left: distributions of (a) clip duration, (b) expression length, (c) target interval duration, (d) interval-to-clip duration ratio, (e) target count, and (f) box count per expression. Panels (c), (d), and (f) cover expressions with matching targets. Dashed lines mark means. Right: video categories (top) and frequent words in referring expressions (bottom).}
  \label{fig:dataset-statistics}
\end{figure*}

\subsection{Data Collection and Annotation}
\label{sec:data-collection-annotation}

\paragraph{Data collection.}
Unlike existing benchmarks such as VidSTG and BOSTVG that rely on shorter videos~\cite{zhang2020does,yao2026omnistvg}, we source long videos from two long-video QA datasets, MLVU dev (569 videos) and LongVideoBench (1,180 videos)~\cite{zhou2025mlvu,wu2024longvideobench}, covering films, documentaries, daily life, and travel with diverse actions and object interactions. Reusing only the raw videos and category metadata, we construct new annotations for referring expressions, target intervals, and frame-level bounding boxes. To prevent repetitive scenes across clips from the same source, we restrict most source videos to 300--1,200 seconds, split them into 300-second segments, and merge shorter remainders into the preceding segment to yield 5--10-minute clips. We retain source mappings, frame timestamps, and clip boundaries for exact annotation alignment, then screen candidate events, construct expressions, and verify temporal uniqueness across the full clip. Dataset statistics are detailed in Sec.~\ref{sec:dataset-statistics} and Supplementary Sec.~A.3.

\paragraph{Zero-target expression construction.}
Easy zero-target queries are constructed by pairing existing expressions with videos whose content differs substantially. Hard zero-target queries are constructed by modifying one key semantic condition in a positive expression for the same video while preserving the remaining scene, action, and relational context as much as possible. Candidates are screened through vision-language model verification and human review.

\subsection{Dataset Characteristics and Quality Analysis}
\label{sec:dataset-statistics}

LV-MOSTG-9K contains 2,369 clips (1,678 from LongVideoBench and 691 from MLVU), each lasting 5--10 minutes (mean: 6.45), totaling 254.777 hours. It comprises 8,788 expressions (5,402 multi-target, 1,373 single-target, and 2,013 zero-target) and 4,423,061 frame-level boxes. Videos span films, documentaries, daily life, and travel; expressions average 21.14 words and describe attributes, actions, and relations. For matching-target expressions, intervals average 12.28 seconds with a median interval-to-video ratio of 1.99\%; each expression refers to 1--12 objects and averages 652.85 boxes, with visible target counts varying across frames. Figure~\ref{fig:dataset-statistics} shows temporal and linguistic distributions (a--d), target counts including zero-target queries (e), box counts for matching-target expressions (f), and video categories and vocabulary (right).

\subsection{Dataset Split and Evaluation Protocol}
\label{sec:evaluation-protocol}

\paragraph{Dataset split.}
The LV-MOSTG-9K test split contains 1,500 expressions. We evaluate grounding on 1,158 expressions with matching targets and zero-target recognition on the full split, which contains 342 zero-target expressions.

We group the 1,158 queries with matching targets by the ratio of the ground-truth interval duration to the full clip duration. The Low group covers $[0\%,1\%)$ (164 queries), Medium covers $[1\%,3\%]$ (483 queries), and High covers $(3\%,100\%]$ (511 queries).

\paragraph{Evaluation metrics.}
Following BOSTVG~\cite{yao2026omnistvg}, we report mean temporal IoU (m-tIoU), mean video IoU (m-vIoU), and vIoU@$R$ ($R\in\{0.3,0.5\}$), aligning predicted and ground-truth box sets via original frame timestamps under BOSTVG's matching and scoring rules (cross-frame instance identities are optional). Inspired by VUE-STG~\cite{josevidi2}, we also report vIoU-Int.\ All on LV-MOSTG-9K: for each query, we average frame-level spatial IoU over the intersection of predicted and ground-truth intervals on the 2~FPS grid, then average over all test expressions with matching targets, assigning zero upon no temporal overlap or spatial failure. For zero-target expressions, where empty predictions are expected, we treat them as the positive class and report precision, recall, and F1.

\section{Long-Video Multi-Object Spatio-Temporal Grounding (LOMOST)}
\label{sec:lomost}

\begin{figure*}[t]
  \centering
  \includegraphics[width=\textwidth]{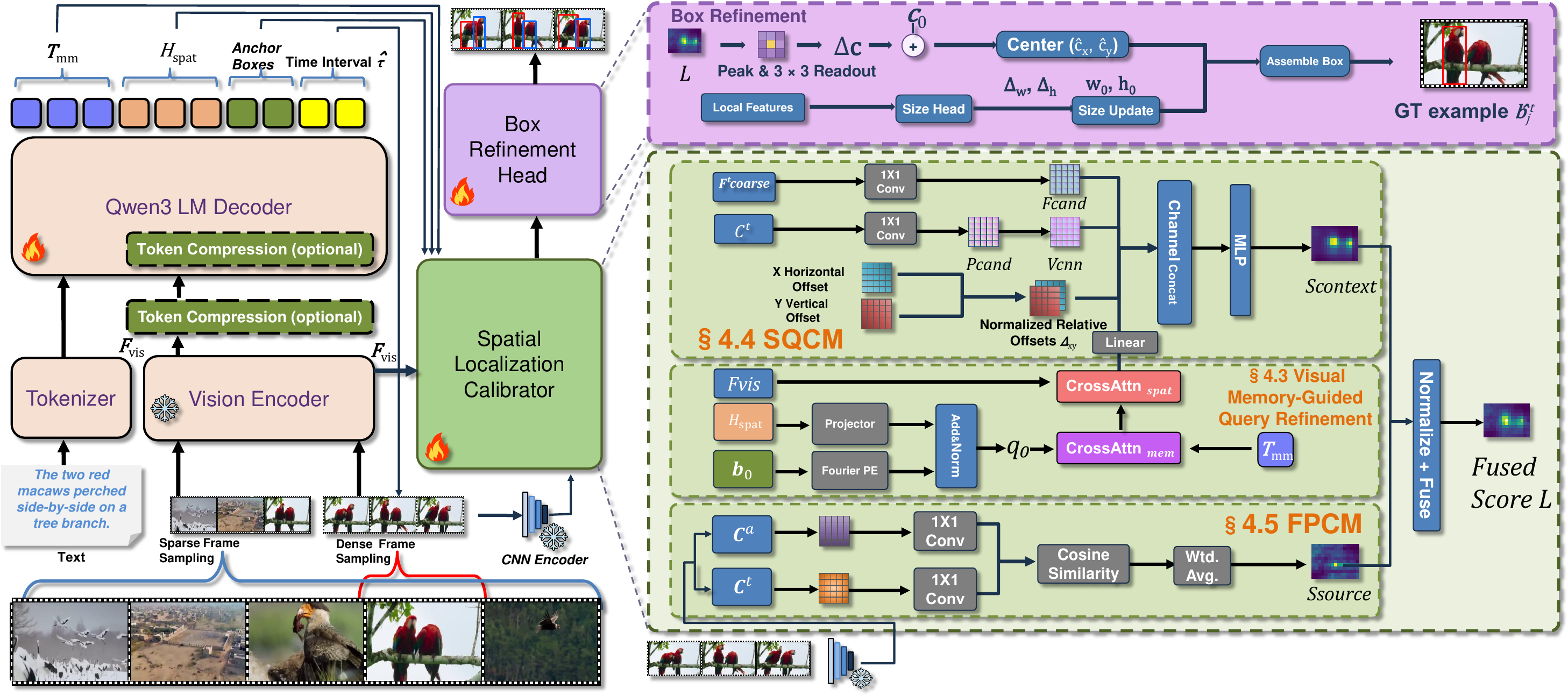}
  \caption{LOMOST architecture. The global stage predicts a target interval and coarse boxes on anchor frames from sparse video observations. The calibrator uses the VLM hidden state associated with each anchor box and the visual memory associated with each anchor frame to refine object queries for frame-level localization. Within the predicted interval, the spatial localization calibrator combines Feature Prior Consistency Modulation (FPCM) and Semantic Query Contextual Modeling (SQCM) to score candidate locations. The box refinement head uses the fused scores and local features to predict frame-level boxes. Dashed compression blocks indicate alternative insertion locations, not sequential stages.}
  \label{fig:lomost}
\end{figure*}

\subsection{Overview}
\label{sec:lomost-overview}
As shown in Fig.~\ref{fig:lomost}, LOMOST combines sparse global understanding with dense frame-level localization.
In the first stage, a supervised fine-tuned Qwen3-VL-4B~\cite{bai2025qwen3} takes sparsely sampled frames from the full video and the referring expression as input. It predicts a target interval $\hat{\tau}=[\hat{s},\hat{e})$ and coarse bounding boxes on three anchor frames. Inspired by the use of token representations to condition dense prediction in Qwen3-VL-Seg~\cite{yao2026qwen3}, we introduce a special token \texttt{<SPAT>} and use its hidden-state vector $H_{\mathrm{spat}}$ as an object representation. Each anchor frame also receives a visual memory $\boldsymbol{T}_{\mathrm{mm}}$ from the visual tokens of the nearest sparse visual unit in time. Visual token compression can optionally be applied before or within the language decoder, depending on the strategy; we evaluate its accuracy--efficiency trade-off in Sec.~\ref{sec:token-compression}.

In the second stage, a Spatial Localization Calibrator adapts the sparse anchor-frame priors to each current frame within $\hat{\tau}$. We extract multilayer visual features $\boldsymbol{F}_{\mathrm{vis}}$ for these frames. Visual Memory-Guided Query Refinement combines the object representation $H_{\mathrm{spat}}$ and coarse box $\boldsymbol{b}_0$ from the selected anchor to initialize an object query $\boldsymbol{q}_0$. Memory cross-attention first updates the query with $\boldsymbol{T}_{\mathrm{mm}}$ to obtain $\boldsymbol{q}_{\mathrm{mm}}$. Spatial cross-attention then incorporates the fused visual features of the current frame to obtain $\boldsymbol{q}_1$.

The calibrator integrates two complementary branches on a shared candidate grid. Semantic Query Contextual Modeling (SQCM) uses the refined object query and current-frame features to evaluate candidate locations. Feature Prior Consistency Modulation (FPCM) supplies anchor-referenced appearance consistency scores. Their scores are fused into a map $\boldsymbol{L}$, combining query-conditioned context with the anchor's appearance prior. The Box Refinement Head reads the center displacement from $\boldsymbol{L}$ and predicts width and height residuals from local features to refine the coarse box, yielding a set of target boxes for each processed frame within $\hat{\tau}$.

\subsection{Sparse Global Understanding}
\label{sec:sparse-global-understanding}

We sparsely sample frames from the full video and feed them, along with their timestamps and the referring expression, to Qwen3-VL-4B. We fine-tune the model to produce structured grounding outputs. To construct the supervision, we use the $1/6$, $1/2$, and $5/6$ positions within the ground-truth target interval as reference points and select three anchor frames from the sparsely sampled frames with valid box annotations. The model learns to predict the target interval $\hat{\tau}$, the three anchor timestamps, and the list of coarse boxes at each anchor, with one \texttt{<SPAT>} token per box. We pair each box with the hidden state of its token to provide geometric and semantic information for the corresponding object query.

The global stage passes the following representation to the frame-level localization stage:
\begin{equation}
  \begin{array}{rl}
    \mathcal{G} &= \left(\hat{\tau},
      \left\{\left(u_k, \mathcal{A}_k, \boldsymbol{T}_{\mathrm{mm}}^{k}\right)\right\}_{k=1}^{3}\right), \\
    \mathcal{A}_k &= \left\{(\boldsymbol{b}_{0,j}^{k}, H_{\mathrm{spat},j}^{k})\right\}_{j=1}^{N_k}.
  \end{array}
\end{equation}
Here, $u_k$ and $N_k$ are the timestamp and number of coarse boxes of anchor $k$. The pair $(\boldsymbol{b}_{0,j}^{k}, H_{\mathrm{spat},j}^{k})$ contains the $j$-th coarse box and the hidden state of its associated \texttt{<SPAT>} token. All object queries from anchor $k$ share $\boldsymbol{T}_{\mathrm{mm}}^{k}$, the visual tokens of the sparse visual unit closest in time to $u_k$.

\subsection{Visual Memory-Guided Query Refinement}
\label{sec:query-refinement}

For each frame at time $t$ in $\hat{\tau}$, we select the temporally nearest anchor $a$ and initialize one query per coarse box $j$:
\begin{equation}
  \boldsymbol{q}_{0,j}=\mathrm{LN}\bigl(
    \boldsymbol{W}_s\mathrm{LN}(H_{\mathrm{spat},j}^{a})
    +\mathrm{MLP}_{\mathrm{box}}(\gamma(\boldsymbol{b}_{0,j}^{a}))
  \bigr).
\end{equation}
Here, $H_{\mathrm{spat},j}^{a}$ is the associated \texttt{<SPAT>} hidden state, and $\gamma$ Fourier-encodes the box center and log width/height; both terms are mapped to the same dimension. We project anchor memory $\boldsymbol{T}_{\mathrm{mm}}^{a}$ and add 1D positional encoding to obtain $\boldsymbol{M}^{a}$. We fuse four layers of current-frame features $\boldsymbol{F}_{\mathrm{vis}}^{t}$ and add 2D positional encoding to obtain $\boldsymbol{F}_{\mathrm{coarse}}^{t}$. Two successive cross-attentions incorporate anchor memory and spatial evidence:
\begin{align}
  \boldsymbol{q}_{\mathrm{mm},j}
    &=\mathrm{CrossAttn}_{\mathrm{mem}}(\boldsymbol{q}_{0,j},\boldsymbol{M}^{a}),\\
  \boldsymbol{q}_{1,j}^{t}
    &=\mathrm{CrossAttn}_{\mathrm{spat}}(\boldsymbol{q}_{\mathrm{mm},j},\boldsymbol{F}_{\mathrm{coarse}}^{t}).
\end{align}
The refined query is passed to SQCM.

\subsection{Semantic Query Contextual Modeling}
\label{sec:context-scoring}

The SQCM branch models current-frame context conditioned on the refined semantic query. For each object query, we construct a $15\times15$ grid of candidate centers around the center of its anchor-frame coarse box $\boldsymbol{b}_0$. The SQCM and FPCM branches share this grid.

We sample the fused current-frame feature map $\boldsymbol{F}_{\mathrm{coarse}}^{t}$ at the candidate centers and apply a $1\times1$ convolution to obtain candidate visual features $\boldsymbol{F}_{\mathrm{cand}}$. We also extract a $7\times7$ patch from the current-frame CNN feature map $\boldsymbol{C}^{t}$ for each candidate center. A shared $1\times1$ convolution projects the CNN features from 128 to 32 channels. We then take a normalized weighted average over the 49 positions in each patch and stack the resulting descriptors over the candidate grid to obtain $\boldsymbol{V}_{\mathrm{cnn}}$.

We project the updated object query $\boldsymbol{q}_{1,j}^{t}$ for object $j$ to 16 dimensions and broadcast it over the $15\times15$ grid to obtain $\boldsymbol{Q}_{\mathrm{grid}}$. For each candidate, we record the horizontal and vertical offsets from the coarse box center and normalize them by the box width and height, respectively. Stacking these offsets over the grid gives $\boldsymbol{\Delta}_{\mathrm{xy}}$. We also include the grid of valid sampling weight fractions, $\boldsymbol{V}_{\mathrm{valid}}$. We concatenate these features along the channel dimension and use a small convolutional network to score the candidates:
\begin{equation}
  \begin{aligned}
    \boldsymbol{X} = \mathrm{Concat}_{\mathrm{ch}}\bigl(
      &\boldsymbol{F}_{\mathrm{cand}}, \boldsymbol{V}_{\mathrm{cnn}}, \boldsymbol{Q}_{\mathrm{grid}},\\
      &\boldsymbol{\Delta}_{\mathrm{xy}}, \boldsymbol{V}_{\mathrm{valid}}\bigr),
  \end{aligned}
\end{equation}
\begin{equation}
  \boldsymbol{S}_{\mathrm{context}} = f_{\mathrm{context}}(\boldsymbol{X}).
\end{equation}
The SQCM branch outputs a $15\times15$ context score map $\boldsymbol{S}_{\mathrm{context}}$ for each object query.

\subsection{Feature Prior Consistency Modulation}
\label{sec:appearance-matching}

The FPCM branch assesses appearance consistency between current-frame candidates and the anchor-frame target. We use the center, width, and height of the coarse box $\boldsymbol{b}_0$ to sample a $7\times7$ appearance template $\boldsymbol{P}_{\mathrm{anc}}^a$ from the anchor-frame CNN feature map $\boldsymbol{C}^a$. We reuse the candidate center grid defined in Sec.~\ref{sec:context-scoring}. For each candidate center $\boldsymbol{c}$, we extract a $7\times7$ feature patch $\boldsymbol{P}_{\mathrm{cand}}^t(\boldsymbol{c})$ centered at $\boldsymbol{c}$ from the current-frame CNN feature map $\boldsymbol{C}^t$, using the same relative sampling layout as the anchor template.

We apply the same learnable $1\times1$ convolution to $\boldsymbol{P}_{\mathrm{anc}}^a$ and $\boldsymbol{P}_{\mathrm{cand}}^t(\boldsymbol{c})$ to project the features at each sampling location to 32 dimensions. We denote the projected features at relative position $r$ by $\boldsymbol{f}_r^a$ for the anchor template and $\boldsymbol{f}_{\boldsymbol{c},r}^t$ for the candidate patch. For each valid candidate center $\boldsymbol{c}$, we compute the cosine similarity at corresponding positions and take a weighted average to obtain its appearance consistency score.
\begin{equation}
  s_{\mathrm{source}}(\boldsymbol{c})
  = \sum_{r=1}^{49} \tilde{w}_{\boldsymbol{c},r}\,
    \operatorname{cos}(\boldsymbol{f}_r^a, \boldsymbol{f}_{\boldsymbol{c},r}^t),
\end{equation}
where $\tilde{w}_{\boldsymbol{c},r}$ is the normalized spatial weight. We arrange the candidate scores on the grid to obtain the $15\times15$ appearance consistency score map $\boldsymbol{S}_{\mathrm{source}}$.

Finally, we normalize the FPCM and SQCM score maps, $\boldsymbol{S}_{\mathrm{source}}$ and $\boldsymbol{S}_{\mathrm{context}}$, separately over valid candidate locations to obtain $\boldsymbol{Z}_{\mathrm{source}}$ and $\boldsymbol{Z}_{\mathrm{context}}$. We combine them using learnable positive weights $\alpha$ and $\beta$ as $\boldsymbol{L}=\alpha \boldsymbol{Z}_{\mathrm{source}}+\beta \boldsymbol{Z}_{\mathrm{context}}$. The fused map $\boldsymbol{L}$ contains the final scores for the 225 candidate displacements. The box refinement head uses this map to read out the target center displacement.

\subsection{Box Refinement Head}
\label{sec:box-prediction-head}

To update the center and dimensions of each coarse box $\boldsymbol{b}_0$, we first add the weighted-average displacement from the $3\times3$ neighborhood of the highest-scoring candidate in $\boldsymbol{L}$ to the center of $\boldsymbol{b}_0$, obtaining the predicted center $(\hat c_x,\hat c_y)$. Using the center, width, and height of $\boldsymbol{b}_0$, we then define a $15\times15$ local grid $\mathcal{G}_0$, on which we sample and project current-frame fused visual and CNN features to obtain $\boldsymbol{F}_{\mathrm{loc}}$ and $\boldsymbol{C}_{\mathrm{loc}}$, and project and broadcast $\boldsymbol{q}_{1,j}^{t}$ to obtain $\boldsymbol{Q}_{\mathrm{loc}}$. We also compute the similarity between $\boldsymbol{q}_{1,j}^{t}$ and $\boldsymbol{F}_{\mathrm{coarse}}^{t}$ at each spatial location $p$ (with feature vector $\boldsymbol{f}_{\mathrm{coarse}}^t(p)$ and projection dimension $d$), sample the normalized response $\boldsymbol{P}_{\mathrm{loc}}$ on $\mathcal{G}_0$, and concatenate it with local features, relative grid coordinates $\boldsymbol{\Delta}_{\mathrm{loc}}$, and validity mask $\boldsymbol{M}_{\mathrm{loc}}$ into $\boldsymbol{X}_{\mathrm{size}}=\allowbreak\operatorname{Concat}_{\mathrm{ch}}(\boldsymbol{F}_{\mathrm{loc}},\allowbreak \boldsymbol{C}_{\mathrm{loc}},\allowbreak \boldsymbol{Q}_{\mathrm{loc}},\allowbreak \boldsymbol{P}_{\mathrm{loc}},\allowbreak \boldsymbol{\Delta}_{\mathrm{loc}},\allowbreak \boldsymbol{M}_{\mathrm{loc}})$ to predict width and height residuals $(\Delta_w,\Delta_h)$ and update $(w_0,h_0)$:
\begin{align}
  \boldsymbol{P}_{\mathrm{loc}}
  &= \boldsymbol{M}_{\mathrm{loc}}\odot\operatorname{Sample}_{\mathcal{G}_0}\!\Bigl[\mathcal{N}_{p}\!\Bigl(
     \nonumber\\[-2pt]
  &\quad\tfrac{(\boldsymbol{W}_q\boldsymbol{q}_{1,j}^{t})^\top
    (\boldsymbol{W}_k\boldsymbol{f}_{\mathrm{coarse}}^{t}(p))}{\sqrt{d}}\Bigr)\Bigr],\\
  (\tilde w,\tilde h)
  &=\sigma\!\Bigl(\operatorname{logit}(w_0,h_0)
    \nonumber\\[-2pt]
  &\quad+\operatorname{MLP}\!\bigl(\operatorname{Flatten}(\operatorname{Conv}(\boldsymbol{X}_{\mathrm{size}}))\bigr)\Bigr),
\end{align}
where $\mathcal{N}_p$ applies spatial softmax over valid locations and divides by the peak response. Finally, we constrain $(\tilde w,\tilde h)$ so that the box at $(\hat c_x,\hat c_y)$ lies within the image boundaries, yielding $\hat{\boldsymbol{b}}_j^t=(\hat c_x,\hat c_y,\hat w,\hat h)$.

\paragraph{Training objectives.}
We first fine-tune the VLM, then freeze it to train the localization modules:
\begin{equation}
  \mathcal{L}=
  \begin{cases}
    \mathcal{L}_{\mathrm{AR}}+\mathcal{L}_{\mathrm{token}}, & \text{stage 1},\\
    \mathcal{L}_{\mathrm{box}}+\mathcal{L}_{\mathrm{cand}}+\mathcal{L}_{\mathrm{aux}}, & \text{stage 2}.
  \end{cases}
  \label{eq:training-objectives}
\end{equation}
Loss definitions, weights, and training details are in Supplementary Secs.~B.1--B.2.

\setlength{\dbltextfloatsep}{10pt plus 2pt minus 2pt}
\setlength{\dblfloatsep}{6pt plus 2pt minus 2pt}
\section{Experiments}
\label{sec:experiments}

\begin{table*}[!t]
  \centering
  \caption{Results on the LV-MOSTG-9K test split (1,500 expressions, \%). Grounding metrics use 1,158 expressions with matching targets. Zero-target recognition treats the 342 no-match expressions as the positive class. For LOMOST, 4B denotes the backbone scale. Dashes indicate unavailable results. Best scores are \textbf{bold} and second-best scores are \underline{underlined}.}
  \label{tab:lv-mostg}
  \footnotesize
  \renewcommand{\arraystretch}{0.88}
  \setlength{\tabcolsep}{1.15pt}
  \begin{tabular}{@{}l*{14}{c}@{}}
    \toprule
    & \multicolumn{5}{c}{Overall grounding}
    & \multicolumn{6}{c}{Grounding by interval ratio}
    & \multicolumn{3}{c}{Zero-target recognition} \\
    \cmidrule(lr){2-6}\cmidrule(lr){7-12}\cmidrule(l){13-15}
    & \multicolumn{5}{c}{} & \multicolumn{2}{c}{Low (164)} & \multicolumn{2}{c}{Medium (483)} & \multicolumn{2}{c}{High (511)} & \multicolumn{3}{c}{} \\
    \cmidrule(lr){7-8}\cmidrule(lr){9-10}\cmidrule(lr){11-12}
    Method & m-tIoU & m-vIoU & vIoU-Int. All & vIoU@0.3 & vIoU@0.5
           & m-tIoU & m-vIoU & m-tIoU & m-vIoU & m-tIoU & m-vIoU & Prec. & Rec. & F1 \\
    \midrule
    TubeDETR (adapted)~\cite{yang2022tubedetr} & 33.42 & 16.48 & 24.77 & 24.44 & 9.59 & 30.35 & 13.95 & 34.48 & 17.30 & 33.40 & 16.51 & 79.84 & 59.06 & 67.90 \\
    STCAT (adapted)~\cite{jin2022embracing}        & 25.93 & 12.08 & 19.07 & 16.75 & 6.22 & 32.71 & 15.07 & 30.18 & 14.20 & 19.73 & 9.11 & 76.71 & 55.85 & 64.64 \\
    TA-STVG (adapted)                       & 34.29 & \underline{23.64} & \underline{33.12} & \underline{38.26} & \underline{21.07} & \underline{34.69} & \textbf{25.05} & 36.41 & \underline{24.76} & 32.16 & \underline{22.12} & 76.49 & 67.54 & 71.74 \\
    \midrule
    GLM-4.6V-Flash (9B)~\cite{hong2025glm} & 34.07 & 8.34 & 14.90 & 3.11 & 0.17 & 24.55 & 6.04 & 34.93 & 8.65 & 36.31 & 8.78 & 83.52 & 66.67 & 74.15 \\
    InternVideo3-8B~\cite{yan2026internvideo3} & 11.29 & 3.05 & 10.03 & 0.60 & 0.09 & 6.78 & 1.35 & 11.22 & 3.00 & 12.80 & 3.64 & 52.91 & 34.50 & 41.77 \\
    Qwen3.5-4B~\cite{qwen3.5} & 3.24 & 0.73 & 3.56 & 0.00 & 0.00 & 0.76 & 0.13 & 1.46 & 0.42 & 5.72 & 1.21 & 49.25 & \underline{85.96} & 62.62 \\
    Qwen3.5-27B~\cite{qwen3.5} & 36.65 & 9.20 & 13.54 & 8.29 & 0.69 & 18.20 & 4.94 & 32.93 & 7.53 & 46.08 & 12.15 & 45.15 & 51.75 & 48.23 \\
    Qwen3.5-122B-A10B~\cite{qwen3.5} & 38.91 & 6.91 & 10.60 & 3.20 & 0.35 & 21.62 & 4.62 & 36.83 & 6.79 & 46.42 & 7.76 & 72.39 & 34.50 & 46.73 \\
    LLaVA-OneVision-2-8B & 3.20 & 0.69 & 4.80 & 0.26 & 0.00 & 1.02 & 0.15 & 3.82 & 0.78 & 3.31 & 0.78 & 34.10 & 39.18 & 36.46 \\
    \midrule
    Gemini 3.5 Flash-Lite~\cite{google2026gemini35flashlite} & 45.84 & 13.34 & 21.91 & 13.64 & 2.25 & 34.57 & 11.56 & \textbf{52.73} & 15.01 & 42.95 & 12.32 & \underline{89.26} & 70.47 & \underline{78.76} \\
    GPT-5.6 Luna~\cite{openai2026gpt56luna} & \underline{47.33} & 14.75 & 23.39 & 13.82 & 1.47 & 31.23 & 9.53 & 46.19 & 14.49 & \textbf{53.59} & 16.67 & 88.79 & 30.12 & 44.98 \\
    \midrule
    LOMOST (4B, ours) & \textbf{49.34} & \textbf{30.27} & \textbf{48.16} & \textbf{49.91} & \textbf{22.11} & \textbf{37.78} & \underline{23.40} & \underline{50.28} & \textbf{31.09} & \underline{52.15} & \textbf{31.70} & \textbf{92.92} & \textbf{92.11} & \textbf{92.51} \\
    \bottomrule
  \end{tabular}
\end{table*}

\begin{table*}[!t]
\begin{minipage}[t]{\columnwidth}
  \centering
  \caption{Results on the official BOSTVG test split (1,912 expressions). Scores are percentages, with failed predictions scored as zero. Best scores are \textbf{bold} and second-best scores are \underline{underlined}.}
  \label{tab:bostvg}
  \small
  \renewcommand{\arraystretch}{0.88}
  \setlength{\tabcolsep}{3pt}
  \begin{tabular}{@{}lrrrr@{}}
    \toprule
    Method & m-tIoU & m-vIoU & vIoU@0.3 & vIoU@0.5 \\
    \midrule
    TubeDETR (adapted) & 31.05 & 7.52 & 3.14 & 0.17 \\
    STCAT (adapted) & 33.31 & 8.03 & 3.35 & 0.31 \\
    CG-STVG (adapted) & 32.09 & 7.66 & 3.56 & 0.26 \\
    BOSTVG baseline & 25.73 & 5.38 & 1.10 & 0.21 \\
    OmniTube & \underline{35.83} & \underline{9.47} & \underline{6.17} & \underline{0.89} \\
    \midrule
    LOMOST (4B, ours) & \textbf{36.03} & \textbf{22.10} & \textbf{34.78} & \textbf{12.50} \\
    \bottomrule
  \end{tabular}
\end{minipage}\hfill
\begin{minipage}[t]{\columnwidth}
  \centering
  \caption{Results on the full VidSTG test split (10,303 expressions). All scores are percentages. Baselines are re-evaluated with their official checkpoints under the common input and scoring protocol. Best scores are \textbf{bold} and second-best scores are \underline{underlined}.}
  \label{tab:vidstg}
  \small
  \renewcommand{\arraystretch}{0.88}
  \setlength{\tabcolsep}{3.5pt}
  \begin{tabular}{@{}lrrrr@{}}
    \toprule
    Method & m-tIoU & m-vIoU & vIoU@0.3 & vIoU@0.5 \\
    \midrule
    TubeDETR~\cite{yang2022tubedetr} & \underline{45.08} & 25.79 & 36.02 & 22.60 \\
    STCAT~\cite{jin2022embracing} & 44.05 & 26.41 & 37.68 & 22.35 \\
    CG-STVG~\cite{gu2024context} & 43.91 & 27.28 & 39.83 & 23.66 \\
    TA-STVG & 44.54 & \underline{27.50} & \underline{40.22} & \underline{24.23} \\
    \midrule
    LOMOST (4B, ours) & \textbf{50.28} & \textbf{28.34} & \textbf{41.33} & \textbf{27.38} \\
    \bottomrule
  \end{tabular}
\end{minipage}
\end{table*}

\subsection{Experimental Setup}
\label{sec:experimental-setup}

\paragraph{Implementation Details.}
We build LOMOST on Qwen3-VL-4B~\cite{bai2025qwen3} using PyTorch~\cite{paszke2019pytorch} and conduct experiments on LV-MOSTG-9K, BOSTVG, and VidSTG. All experiments run on a server equipped with four NVIDIA A800 GPUs.

\paragraph{Comparison Protocol.}
We compare against task-specific STVG models (TubeDETR, STCAT, CG-STVG, TA-STVG, and OmniTube), open-source VLMs (GLM, InternVideo3, Qwen3.5, and LLaVA-OneVision-2), and proprietary API models (Gemini and GPT). Tables~\ref{tab:lv-mostg}--\ref{tab:vidstg} list dataset-specific baselines and model variants.
\textbf{LV-MOSTG-9K.} Adapted baselines follow Sec.~\ref{sec:evaluation-protocol}, with spatial scoring at 2~FPS.
\textbf{BOSTVG.} Baseline results follow OmniSTVG~\cite{yao2026omnistvg}; LOMOST uses the official training split and the same test protocol with native-timestamp scoring. On both datasets, LOMOST uses 0.5/2~FPS for global/local processing during training and inference.
\textbf{VidSTG.} We re-evaluate four official baseline checkpoints on 10,303 test expressions using shared 2~FPS inputs and uncapped 2~FPS scoring. LOMOST uses 64/128-frame caps for training/evaluation and localizes within predicted intervals. Box interpolation and further settings are in Supplementary Secs.~C.1--C.3.

\subsection{Main Results}
\label{sec:main-results}
\paragraph{LV-MOSTG-9K.}
LOMOST (4B) outperforms all evaluated open-source vision-language baselines in overall spatio-temporal grounding (Tab.~\ref{tab:lv-mostg}). Compared with TA-STVG, the strongest task-specific baseline, LOMOST achieves 30.27\% m-vIoU and 48.16\% vIoU-Int.\ All, gains of 6.63 and 15.04 percentage points, respectively. LOMOST also achieves the highest overall m-tIoU (49.34\%), 2.01 points above GPT-5.6 Luna, the strongest baseline on this metric. LOMOST reaches 92.51\% zero-target F1, 13.75 points above Gemini 3.5 Flash-Lite, the strongest baseline on this metric. In the Low interval-ratio group, LOMOST exceeds TA-STVG in m-tIoU (37.78\% vs.\ 34.69\%) but trails in m-vIoU (23.40\% vs.\ 25.05\%). Global sampling at 0.5~FPS may miss spatial evidence in brief events despite better temporal overlap. For comparison with a task-fine-tuned VLM of the same size, we report Qwen3-VL-4B with nearest-anchor box copying in Table~\ref{tab:ablation} and with linear interpolation in Supplementary Table~\ref{tab:supp-sft-linear}.

\paragraph{BOSTVG.}
LOMOST achieves 22.10\% m-vIoU on the official test split, improving on OmniTube by 12.63 percentage points (Tab.~\ref{tab:bostvg}).

\paragraph{VidSTG.}
On this single-object grounding benchmark, LOMOST leads the evaluated methods on all four reported metrics under the common evaluation protocol (Tab.~\ref{tab:vidstg}). It reaches 28.34\% m-vIoU and improves m-tIoU and m-vIoU by 5.20 and 0.84 percentage points over the strongest baseline for each metric, respectively.

\subsection{Ablation Studies}
\label{sec:ablation-studies}

\paragraph{Ablation Settings.}
Table~\ref{tab:ablation} compares four configurations on BOSTVG TEST1912 and LV-MOSTG-9K TEST1500.
\textbf{(1) Qwen + Box Copy} uses zero-shot Qwen3-VL-4B and copies boxes from the nearest anchor frame to each frame in the predicted interval.
\textbf{(2) SFT + Box Copy} applies the same copying rule to the predictions of the task-fine-tuned Qwen3-VL-4B, allowing a direct comparison before and after supervised fine-tuning.
Configurations (2)--(4) share the same fine-tuned VLM, predicted intervals, and anchor boxes within each benchmark.
\textbf{(3) SFT + FPCM} uses FPCM alone, while \textbf{(4) SFT + FPCM + SQCM (LOMOST)} uses both modules.

\paragraph{Results and Analysis.}
Compared with SFT + Box Copy, LOMOST with FPCM and SQCM improves m-vIoU by 1.54 points on BOSTVG and 1.60 points on LV-MOSTG-9K, and improves vIoU-Int.\ All by 3.13 and 2.27 points, respectively (Tab.~\ref{tab:ablation}). On LV-MOSTG-9K, zero-target precision, recall, and F1 are 97.08\%, 38.89\%, and 55.53\% for the original Qwen and 92.92\%, 92.11\%, and 92.51\% for all three SFT configurations.

\begin{table}[!htbp]
  \centering
  \caption{Ablations on BOSTVG TEST1912 and LV-MOSTG-9K TEST1500 (\%). Parentheses show changes in percentage points over the previous configuration on the same split. vIoU-Int.\ All uses per-query vIoU/tIoU on BOSTVG (zero for no overlap) and the definition in Sec.~\ref{sec:evaluation-protocol} on LV-MOSTG-9K.}
  \label{tab:ablation}
  \scriptsize
  \setlength{\tabcolsep}{1.5pt}
  \newcommand{\ablationgain}[2]{#1\kern0.5pt{\fontsize{5.5}{6.5}\selectfont (#2)}}
  \resizebox{\linewidth}{!}{%
  \begin{tabular}{@{}lccccc@{}}
    \toprule
    Configuration & m-tIoU & m-vIoU & Int.\ All & @0.3 & @0.5 \\
    \midrule
    \multicolumn{6}{@{}l}{\textit{BOSTVG TEST1912}} \\
    Qwen + Box Copy & 22.64 & 2.08 & 8.96 & 0.21 & 0.00 \\
    SFT + Box Copy & \ablationgain{36.03}{+13.39} & \ablationgain{20.56}{+18.48} & \ablationgain{42.00}{+33.04} & \ablationgain{30.13}{+29.92} & \ablationgain{10.51}{+10.51} \\
    SFT + FPCM & \ablationgain{36.03}{0.00} & \ablationgain{21.54}{+0.98} & \ablationgain{43.92}{+1.92} & \ablationgain{32.69}{+2.56} & \ablationgain{11.72}{+1.21} \\
    \begin{tabular}[c]{@{}l@{}}SFT + FPCM + SQCM\\(LOMOST)\end{tabular} & \ablationgain{36.03}{0.00} & \ablationgain{22.10}{+0.56} & \ablationgain{45.13}{+1.21} & \ablationgain{34.78}{+2.09} & \ablationgain{12.50}{+0.78} \\
    \midrule
    \multicolumn{6}{@{}l}{\textit{LV-MOSTG-9K TEST1500}} \\
    Qwen + Box Copy & 7.09 & 0.91 & 4.97 & 0.00 & 0.00 \\
    SFT + Box Copy & \ablationgain{49.34}{+42.25} & \ablationgain{28.67}{+27.76} & \ablationgain{45.89}{+40.92} & \ablationgain{46.20}{+46.20} & \ablationgain{20.29}{+20.29} \\
    SFT + FPCM & \ablationgain{49.34}{0.00} & \ablationgain{29.77}{+1.10} & \ablationgain{46.75}{+0.86} & \ablationgain{47.84}{+1.64} & \ablationgain{21.93}{+1.64} \\
    \begin{tabular}[c]{@{}l@{}}SFT + FPCM + SQCM\\(LOMOST)\end{tabular} & \ablationgain{49.34}{0.00} & \ablationgain{30.27}{+0.50} & \ablationgain{48.16}{+1.41} & \ablationgain{49.91}{+2.07} & \ablationgain{22.11}{+0.18} \\
    \bottomrule
  \end{tabular}%
  }
\end{table}

\subsection{Efficiency Analysis}
\label{sec:efficiency-analysis}

\begin{table}[!t]
  \centering
  \caption{End-to-end latency and peak allocated GPU memory on 32 shared LV-MOSTG-9K queries using one A800 and a query batch size of one. Memory is the maximum per-query peak across the 32 queries.}
  \label{tab:inference-efficiency}
  \small
  \renewcommand{\arraystretch}{0.88}
  \setlength{\tabcolsep}{4pt}
  \begin{tabular}{@{}lrrr@{}}
    \toprule
    Method & Mean (s) & P95 (s) & Peak alloc. (GiB) \\
    \midrule
    TubeDETR & 31.93 & 82.06 & 5.15 \\
    STCAT & 33.06 & 82.34 & 6.95 \\
    TA-STVG & 42.33 & 83.53 & 25.41 \\
    CG-STVG & 77.98 & 111.82 & 25.40 \\
    \midrule
    LOMOST (4B, ours) & 45.35 & 110.55 & 19.83 \\
    \bottomrule
  \end{tabular}
\end{table}

As shown in Tab.~\ref{tab:inference-efficiency}, LOMOST uses 19.83~GiB of peak allocated GPU memory on an A800, compared with 5.15, 6.95, 25.41, and 25.40~GiB for TubeDETR, STCAT, TA-STVG, and CG-STVG, respectively. Its mean end-to-end latency is 45.35~s, exceeding the first three methods by 3.02--13.42~s and falling below CG-STVG's 77.98~s. At this computational cost, LOMOST achieves substantially higher spatio-temporal grounding accuracy (Tab.~\ref{tab:lv-mostg}). Efficiency settings and runtime breakdowns are in Supplementary Sec.~D.1.

\subsection{Visual Token Compression}
\label{sec:token-compression}

We compare five visual token compression strategies at retention rates of 10\%, 20\%, and 50\% to evaluate their effects on long-video spatio-temporal grounding and inference cost. As shown in Tab.~\ref{tab:token-compression}, HoliTom achieves the highest m-vIoU at 10\% and 20\% retention. Its spatio-temporal merging mechanism may be well suited to reducing repeated observations across long background segments~\cite{shao2505holitom}. SemVID achieves the highest m-tIoU at 10\% and 50\%, suggesting that allocating token budgets according to query relevance and inter-frame variation may help preserve boundary cues for brief events~\cite{li2026keeping}. For multi-target expressions, compression should also consider evidence coverage across instances to reduce the risk of discarding briefly visible or less visually salient targets.



\section{Conclusion}
\label{sec:conclusion}
We introduced GL-STVG, a task for generalized spatio-temporal grounding in long videos, along with the LV-MOSTG-9K benchmark and the LOMOST model. To our knowledge, LV-MOSTG-9K is the first STVG benchmark to jointly evaluate zero, single, and multiple target expressions in 5--10-minute videos, with 4.42 million bounding boxes annotated at the original frame rate across 2,369 clips. In LOMOST, SQCM models query-conditioned spatial context to evaluate candidate locations in the current frame, while FPCM computes candidate responses using a cross-frame feature consistency prior derived from the anchor frame. The fused responses guide frame-level box refinement within the predicted interval. Evaluated on three benchmarks, LOMOST achieves state-of-the-art results on BOSTVG and LV-MOSTG-9K.

\begin{table}[!t]
  \centering
  \caption{Token compression on TEST1500 (accuracy: \%). Triplets: 10/20/50\% retention; FULL: 100\% (accuracy from Tab.~\ref{tab:lv-mostg}). Efficiency: 32 shared queries, one A800, batch 1, XGrammar. Bold/underlined: best/second-best per retention rate.}
  \label{tab:token-compression}
  \fontsize{6.5}{7.2}\selectfont
  \setlength{\tabcolsep}{0.5pt}
  \renewcommand{\arraystretch}{1.05}
  \newcommand{\retentiontriplet}[3]{#1/#2/#3}
  \begin{tabular*}{\linewidth}{@{}l@{\hspace{2pt}}@{\extracolsep{\fill}}ccccc@{}}
    \toprule
    Method & m-tIoU $\uparrow$ & m-vIoU $\uparrow$
    & \shortstack{Mean (s)\\$\downarrow$} & \shortstack{P95 (s)\\$\downarrow$}
    & \shortstack{Peak alloc.\\(GiB) $\downarrow$} \\
    \midrule
    FULL & 49.34 & 30.27 & 25.5 & 50.7 & 19.4 \\
    Uniform & \retentiontriplet{29.6}{36.8}{\underline{46.8}} & \retentiontriplet{11.2}{15.0}{25.4} & \retentiontriplet{\textbf{18.5}}{\textbf{21.2}}{\textbf{22.0}} & \retentiontriplet{44.5}{43.6}{\textbf{42.3}} & \retentiontriplet{\textbf{15.4}}{\textbf{15.4}}{\textbf{15.9}} \\
    DivPrune & \retentiontriplet{25.0}{34.0}{45.3} & \retentiontriplet{8.4}{13.7}{23.5} & \retentiontriplet{\underline{20.0}}{22.8}{25.7} & \retentiontriplet{\underline{39.9}}{48.4}{52.2} & \retentiontriplet{\underline{23.8}}{\underline{23.9}}{23.9} \\
    SemVID & \retentiontriplet{\textbf{37.6}}{\underline{42.5}}{\textbf{47.3}} & \retentiontriplet{\underline{14.9}}{\underline{19.9}}{\underline{26.5}} & \retentiontriplet{21.8}{\underline{22.5}}{26.3} & \retentiontriplet{\textbf{39.2}}{\underline{42.2}}{51.3} & \retentiontriplet{\textbf{15.4}}{\textbf{15.4}}{\textbf{15.9}} \\
    SparseVLM+ & \retentiontriplet{21.5}{35.6}{46.6} & \retentiontriplet{8.2}{16.9}{\textbf{26.9}} & \retentiontriplet{20.3}{23.0}{25.7} & \retentiontriplet{48.3}{\textbf{40.0}}{49.6} & \retentiontriplet{42.3}{42.3}{42.9} \\
    HoliTom & \retentiontriplet{\underline{37.5}}{\textbf{45.0}}{46.7} & \retentiontriplet{\textbf{15.5}}{\textbf{23.0}}{\underline{26.5}} & \retentiontriplet{23.9}{24.0}{\underline{25.4}} & \retentiontriplet{55.9}{43.1}{\underline{45.0}} & \retentiontriplet{\textbf{15.4}}{\textbf{15.4}}{\underline{16.0}} \\
    \bottomrule
  \end{tabular*}
\end{table}

\section*{Acknowledgments}
This work was supported in part by the National Natural Science Foundation of China under Grants 61402016 and 62602315, and in part by the Major Key Project of PCL under Grant PCL2025A14.

\section*{LLM Usage Statement}
The authors wrote the initial manuscript draft and used OpenAI GPT for language polishing. The authors selected and verified the references and prepared and checked the figures and tables. OpenAI Codex assisted with writing and reviewing research code under author supervision. We used Qwen to assist with data annotation for LV-MOSTG-9K. The authors designed the experiments, interpreted the results, and developed the research ideas, hypotheses, claims, and conclusions.

{
    \small
    \setlength{\bibsep}{0pt}
    \bibliographystyle{ieeenat_fullname}
    \bibliography{main}
}

\providecommand{\suppmaincite}[1]{\cite{#1}}
\clearpage
\appendix
\twocolumn[{
  \centering
  {\Large\bfseries \thetitle\par}
  \vspace{0.5em}
  {\Large Supplementary Material\par}
  \smallskip
  {\normalsize This supplement describes the construction of LV-MOSTG-9K and presents implementation details and further analyses of LOMOST, our strong baseline for generalized spatio-temporal grounding in long videos (GL-STVG).\par}
  \medskip
}]
\setcounter{dbltopnumber}{3}
\renewcommand{\dblfloatpagefraction}{0.86}
\renewcommand{\dbltopfraction}{0.95}
\renewcommand{\textfraction}{0.04}
\raggedbottom
\setlength{\dbltextfloatsep}{9pt plus 2pt minus 2pt}
\setlength{\dblfloatsep}{8pt plus 2pt minus 2pt}
\setlength{\textfloatsep}{9pt plus 2pt minus 2pt}
\setlength{\floatsep}{8pt plus 2pt minus 2pt}
\setlength{\intextsep}{8pt plus 2pt minus 2pt}

\begingroup
\small
\noindent\textbf{Contents}\par\nobreak
\newcommand{\suppcontentsentry}[2]{%
  \noindent\hyperref[#1]{\ref*{#1}.\enspace#2\dotfill\pageref*{#1}}\par}
\suppcontentsentry{supp:dataset}{Dataset Construction and Analysis}
\suppcontentsentry{supp:training}{Training Objectives and Implementation}
\suppcontentsentry{supp:evaluation}{Evaluation Protocols and Baseline Adaptation}
\suppcontentsentry{supp:efficiency-compression}{Efficiency and Visual Token Compression}
\suppcontentsentry{supp:additional-analysis}{Additional Analysis}
\endgroup

\section{Dataset Construction and Analysis}
\label{supp:dataset}
\subsection{Video Sources and Clip Preparation}
\label{supp:video-preparation}
All LV-MOSTG-9K experiments reported in the main paper use the final version of the benchmark described here, which comprises 2,369 clips. It includes 594 clips from 427 MLVU Dev source videos, 97 clips from 74 MLVU Test source videos, and 1,678 clips from 1,022 LongVideoBench source videos. The source-video counts in Sec.~3.2 refer to the videos selected for the original dataset construction. After further screening and the inclusion of additional samples that passed quality checks, the final benchmark has the source-video and clip counts given above.

For most sources, we select videos between 300 and 1,200 seconds, cut them at frame-aligned five-minute boundaries, and merge any shorter remainder into the preceding clip. For longer sources, we retain up to three non-overlapping windows of approximately five to ten minutes. We retain source identifiers, original frame bounds, and presentation timestamps to map clip annotations back to the source timeline. We screened the clips for shared footage using perceptual frame hashes and manually reviewed the resulting candidates. The final train, validation, and test splits share no source identifiers, exact source-file hashes, or exact clip hashes. 

\subsection{Annotation and Quality Control}
\label{supp:annotation}
\begin{figure}[!htbp]
    \centering
    \includegraphics[width=\columnwidth]{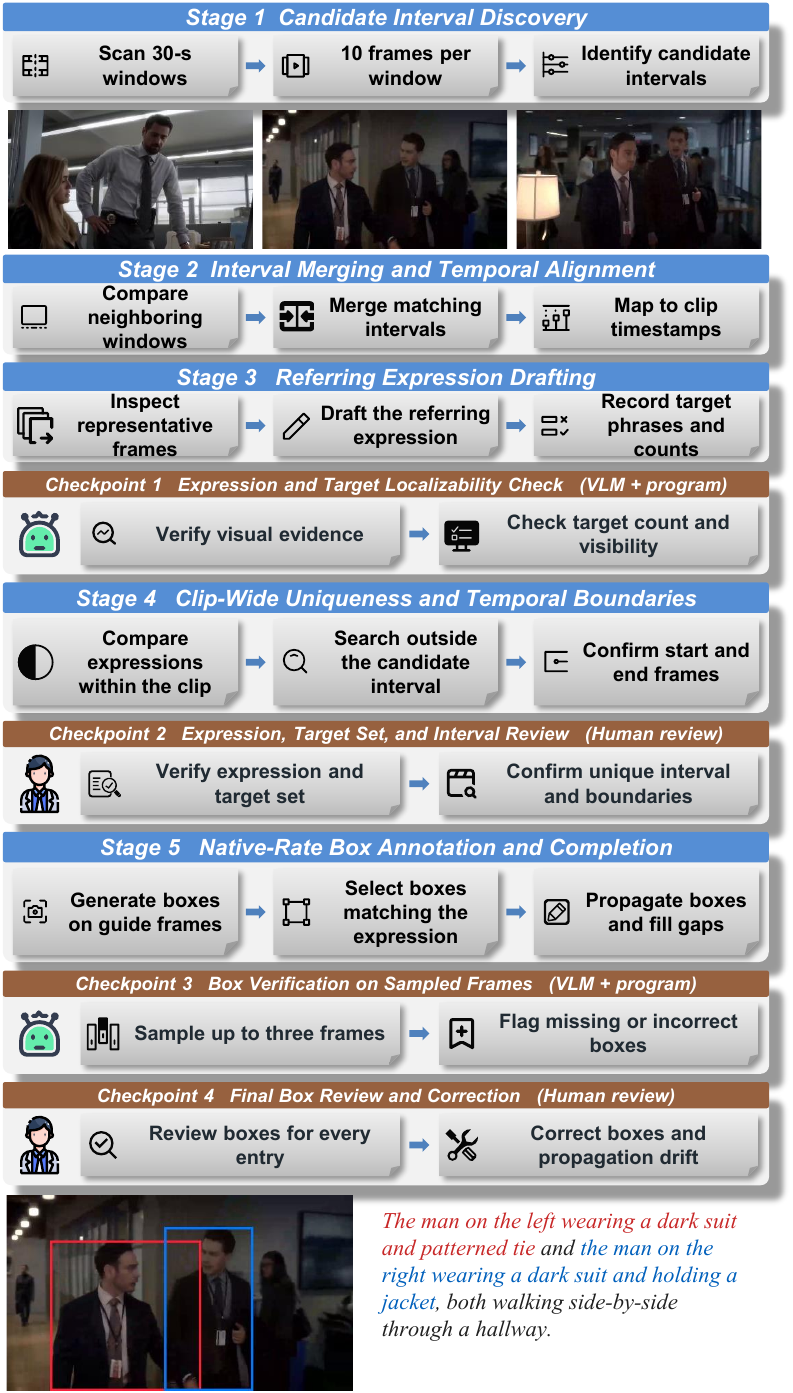}
    \caption{Five annotation stages and four quality checkpoints. Automated checks inspect entries at checkpoints 1 and 3. Human reviewers verify each entry at checkpoints 2 and 4.}
    \label{fig:supp-annotation-pipeline}
\end{figure}

\paragraph{Five-stage annotation pipeline.}
We construct referring expressions with matching targets and their annotations in five stages (Fig.~\ref{fig:supp-annotation-pipeline}). We identify candidate intervals in each full input clip, determine the referring expression, target set, and target time interval, and annotate bounding boxes for the visible targets in each frame of that interval.

\begin{table*}[!tp]
    \centering
    \caption{Counts across annotation and quality-control stages. Each row states its counting unit. Counts from different units or review stages do not form a single retention series.}
    \label{tab:supp-annotation-counts}
    \small
    \begin{tabular}{@{}p{0.29\textwidth}p{0.18\textwidth}p{0.48\textwidth}@{}}
        \toprule
        Stage & Counting unit & Recorded outcome \\
        \midrule
        Candidate interval discovery & Clips, windows, records & 3,834 clips scanned across 50,807 windows, yielding 62,994 valid candidate records. \\
        Candidate interval merging & Groups, records & 5,721 of 7,245 candidate groups merged. 5,733 redundant records skipped. \\
        Expression drafting and Checkpoint 1 & Candidate records & 29,770 of 62,994 passed. Of the remainder, 27,286 were rejected, 5,733 were skipped as redundant, and 205 had processing errors. \\
        Clip-wide uniqueness screening & Expressions & 14,805 of 29,770 retained by automated screening. 11,950 were assigned to human review after further selection. \\
        Checkpoint 2 & Reviewed entries & All 11,950 assigned entries received a decision. Reviewers accepted 9,093, rejected 2,583 as unsuitable, and rejected 274 for repeated matches. \\
        Spatial annotation and box completion & Expressions & 8,932 completed spatial processing. Of the remaining accepted entries, 20 had recorded failures and 141 could not be matched to downstream records. \\
        Checkpoint 3 & Expressions & All 8,932 entries were processed. The VLM reviewed sampled frames for 8,639 entries, and 293 were flagged for manual box annotation. \\
        Checkpoint 4 and initial export & Expressions & 6,872 passed human review, including 2,080 with recorded manual corrections. Of those approved, 6,868 passed the export checks. \\
        \midrule
        Final benchmark & Expressions & 5,468 directly annotated expressions and 1,307 derived single-target expressions with matching targets. The zero-target set contains 1,166 easy and 847 hard expressions. \\
        \bottomrule
    \end{tabular}
\end{table*}

\begin{enumerate}
    \item \textbf{Candidate interval discovery.} We scan each full input clip using windows of approximately 30 seconds and sample 10 frames per window. We prompt Qwen3.7-Flash to identify candidate intervals from the sampled frames, prioritizing intervals with clearly visible targets that can be localized individually. The model returns structured records containing visual descriptions, temporal cues, candidate target categories, and supporting frames.

    \item \textbf{Candidate interval merging and temporal alignment.} Adjacent windows may yield separate records for the same continuous content. Qwen3.7-Flash compares the supporting frames in the two records and the intervening footage to determine whether they refer to the same content. The pipeline merges matching records and their time ranges, retains the correspondence between the original and merged records, and converts times relative to each window to timestamps on the full input clip's timeline.

    \item \textbf{Referring expression drafting and localizability checks.} A vision-language model (VLM) drafts a referring expression based on each candidate interval and its representative frames, using observable attributes, actions, or relationships to distinguish instances of the same category. It records the target phrases and the number of instances each phrase refers to. For example, in \emph{the person standing next to the car}, the car specifies the person's location and the person is the target. At \textbf{Checkpoint 1}, the VLM and automated checks verify that the footage supports the expression, that each target is sufficiently clear, and that each target can be boxed individually.

    \item \textbf{Clip-wide uniqueness verification and temporal boundary confirmation.} We first compare candidate expressions within the same clip, then sample frames from windows outside the candidate interval to check whether another part of the clip satisfies all conditions in the same expression. At \textbf{Checkpoint 2}, human reviewers use the full input clip and the potential matching intervals flagged by the model to confirm the referring expression, the target set, and the start and end frames of the target time interval. Reviewers may revise the expression or the temporal boundaries. If they revise the expression, they reconfirm the target set. We discard entries that still have a distinct match elsewhere in the clip or cannot be annotated reliably.

    \item \textbf{Spatial annotation and box completion at the native frame rate.} For entries that pass Checkpoint 2, the pipeline selects guide frames shot by shot within the target time interval, and SAM3.1 generates candidate boxes. Given the full referring expression, the VLM selects matching instances from the numbered candidate boxes and links each selected box to the corresponding target phrase. SAM3.1 then propagates annotations within local windows in the same shot. The pipeline rechecks portions of the clip with potentially missing boxes and adds boxes for confirmed visible targets. At \textbf{Checkpoint 3}, each entry undergoes automated spatial review, with the VLM inspecting sampled frames when boxes are available. At \textbf{Checkpoint 4}, human reviewers inspect the spatial annotations and correct missing, incorrect, or duplicate boxes, as well as propagation drift. Each frame is annotated only for targets that are visible and identifiable in that frame. If all targets are temporarily invisible in a frame within the target time interval, the frame remains in the interval and its box set is empty.
\end{enumerate}

\begin{figure*}[!tp]
    \centering
    \includegraphics[page=1,width=\textwidth,height=0.94\textheight,keepaspectratio]{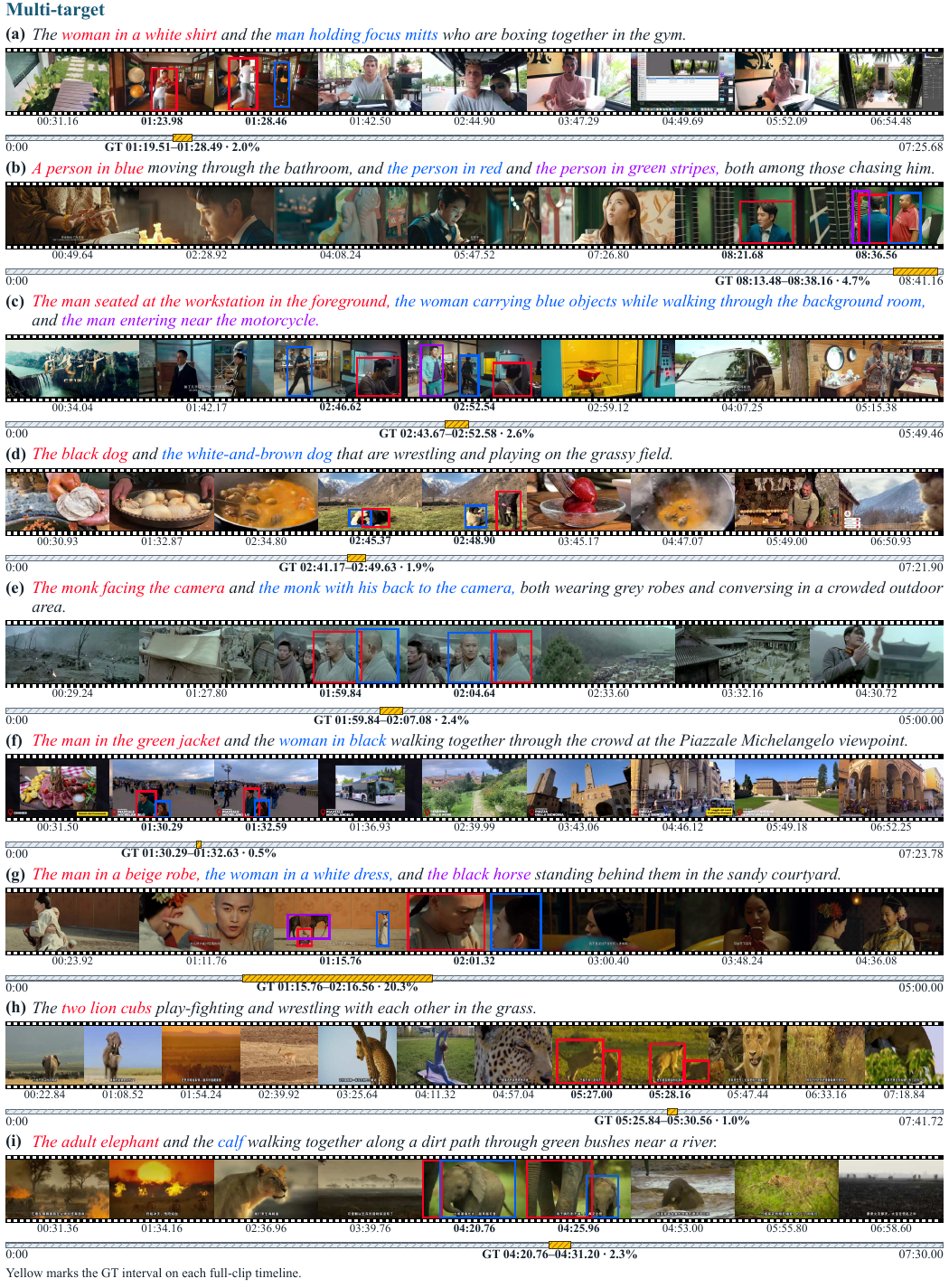}
    \caption{Examples of multi-target referring expressions in LV-MOSTG-9K.}
    \label{fig:supp-annotation-examples-multi}
\end{figure*}

\begin{figure*}[!tp]
    \centering
    \includegraphics[page=2,width=\textwidth,height=0.94\textheight,keepaspectratio]{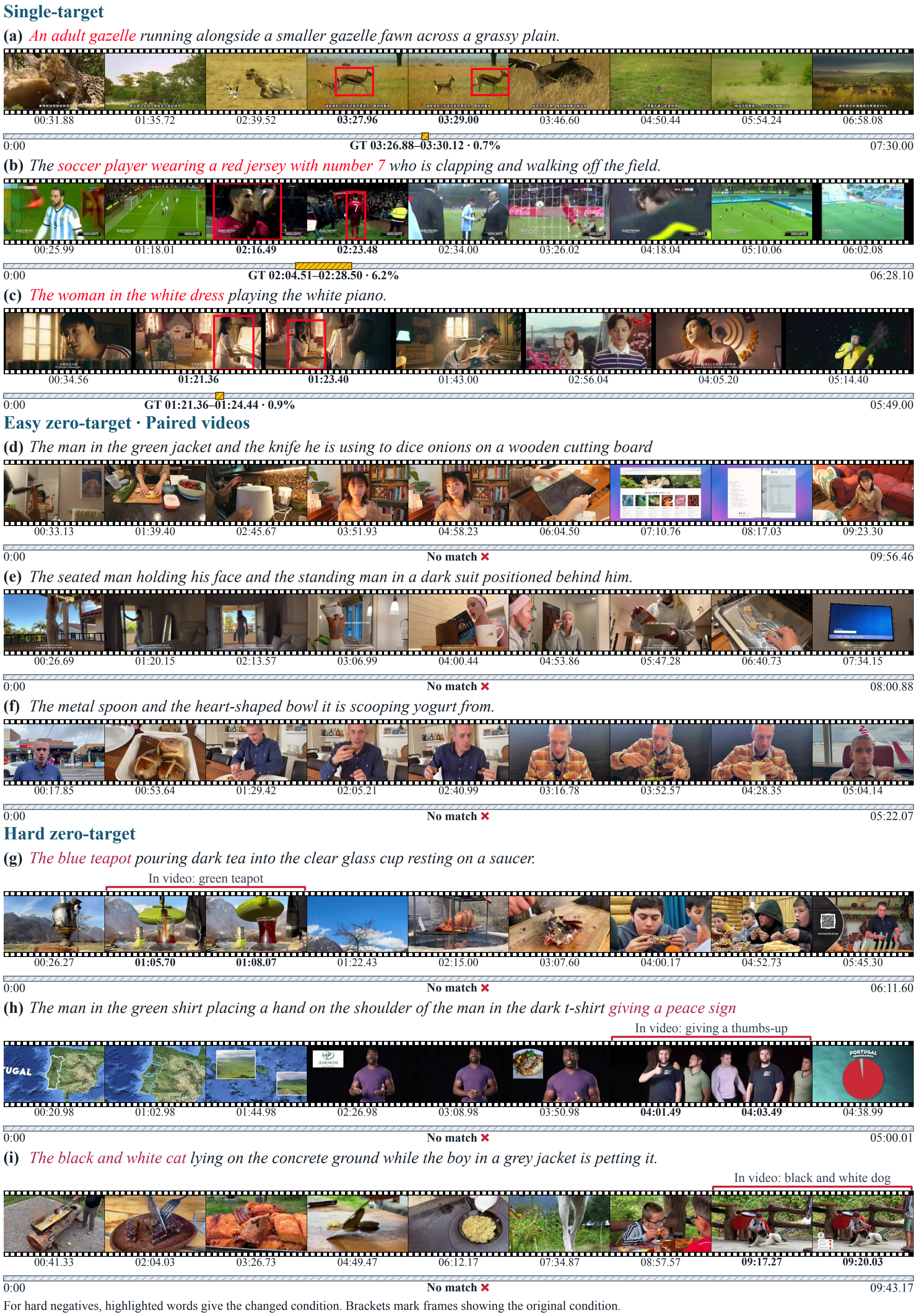}
    \caption{Examples of single-target, easy zero-target, and hard zero-target referring expressions in LV-MOSTG-9K.}
    \label{fig:supp-annotation-examples-single-zero}
\end{figure*}

\begin{table}[!t]
    \centering
    \caption{Video sources by split. Entries give unique source videos / clips. MLVU counts combine its Dev and Test subsets.}
    \label{tab:supp-split-sources}
    \small
    \setlength{\tabcolsep}{3pt}
    \begin{tabular*}{\columnwidth}{@{\extracolsep{\fill}}lrrr@{}}
        \toprule
        Split & MLVU & LongVideoBench & All sources \\
        \midrule
        Train      & 269 / 385 & 591 / 1,005   & 860 / 1,390 \\
        Validation & 59 / 73   & 131 / 183     & 190 / 256 \\
        Test       & 173 / 233 & 300 / 490     & 473 / 723 \\
        \midrule
        Total      & 501 / 691 & 1,022 / 1,678 & 1,523 / 2,369 \\
        \bottomrule
    \end{tabular*}
\end{table}

\paragraph{Quality control.}
Figure~\ref{fig:supp-annotation-pipeline} shows where the four quality checkpoints occur in the five-stage annotation pipeline. A vision-language model (VLM) and automated checks handle Checkpoints 1 and 3. Human reviewers examine every entry assigned to Checkpoints 2 and 4.

\begin{enumerate}
    \item \textbf{Checkpoint 1: Target localizability and expression screening.} We first discard candidates with more than 12 targets. Qwen3.7-Flash then examines representative frames sampled from the candidate interval. It checks whether the target count is manageable, whether the targets are large and clear enough to localize, whether the expression relies on excessively subtle visual details, and whether the footage provides clear evidence for the description. A candidate must pass all four checks to proceed.

    \item \textbf{Checkpoint 2: Human verification and target time interval confirmation.} For every entry assigned to human review, reviewers consult the full input clip and the potential matching intervals flagged by the model. They verify that the targets are distinguishable and that the referring expression corresponds to a single interval that can be annotated reliably. Reviewers may revise the expression, adjust the targets and their counts, and refine the start and end frames of the target time interval. If they revise the expression, they reconfirm the target set. Entries with a distinct match elsewhere in the clip or with targets that cannot be annotated reliably are discarded.

    \item \textbf{Checkpoint 3: Spatial annotation verification.} Every entry reaching this checkpoint is processed. Entries without boxes are flagged for manual box annotation. For each remaining entry, we select up to three representative frames from the beginning, middle, and end of the sequence of frames that have annotated boxes within the target time interval. We overlay the boxes on the selected frames. Given the referring expression, the VLM checks whether any targets lack boxes, whether any boxes mark incorrect objects, and whether the annotated instances match the expression. Its preliminary assessment informs the final human review.

    \item \textbf{Checkpoint 4: Final human review of spatial annotations.} Reviewers inspect the annotations within the target time interval for missing, incorrect, or duplicate boxes and for propagation drift. The interface lets reviewers add, remove, or adjust boxes and propagate annotations again over a specified range using manually drawn guide boxes. Annotations enter the final export only after reviewers confirm that they can be corrected and verified reliably.
\end{enumerate}

\paragraph{Single-target expression construction.}
To help the model learn which targets to localize and how many based on the language of a referring expression, we construct most single-target referring expressions by rewriting verified multi-target expressions. A small number are constructed directly through the five-stage annotation pipeline. The original multi-target expression and the rewritten single-target expression share the same video clip and target time interval, and the rewrite preserves the relevant scene, actions, and interactions as much as possible. The original expression requires localizing multiple objects, and the rewritten expression requires localizing only one of them. The other objects can still serve as references in the description. Compared with using the five-stage pipeline to construct unrelated multi-target and single-target samples separately, this approach holds the visual content fixed, requiring the model to use the constraints in the referring expression to determine which targets to localize and how many. This provides direct supervision for learning target selection and counting. We describe the derivation steps and quality control below.

\begin{enumerate}
    \item \textbf{Screening source samples and candidate targets.} The program first applies strict screening criteria to all source samples. It prioritizes multi-target expressions in which each target phrase refers to exactly one instance. Each target in these expressions has exactly one valid box in every native frame of the original ground-truth interval. This criterion excludes expressions that group multiple instances of the same category under a single plural phrase, such as \emph{two men}. Splitting such a phrase could introduce unverified qualifiers and change the meaning. For eligible targets, the expression can be faithfully rewritten by narrowing its grammatical focus and retaining the other participants as contextual qualifiers.

    \item \textbf{Selecting a target from the video and rewriting the expression.} The rewrite retains one instance as the localization target and preserves the other participants and their relationships with that instance as qualifiers. We call the Qwen3.8-Flash API and uniformly sample the original ground-truth interval at approximately 4 FPS, retaining the first and last annotated frames. If this yields more than 32 frames, we uniformly sample 32 frames across the interval. We provide the VLM with the original expression, the eligible target phrases, and the frames overlaid with the corresponding target IDs. The prompt asks the model to compare the candidates in terms of visual clarity, distinguishability, occlusion, whether the boxes correspond to the objects, and whether the objects can be reliably localized. The model then selects one eligible target and rewrites the expression so that this instance is its sole referent. The prompt instructs it to retain the other participants as context or qualifiers and to preserve the original actions, attributes, temporal order, and roles in the interactions.

    \item \textbf{Retaining the target time interval and the selected target's per-frame boxes.} Given the structured output from Qwen3.8-Flash, the program retains the original video's target time interval and the per-frame boxes for the selected instance. It preserves the box coordinates and records the original frame, the box's position in the original annotation, and the hash.
\end{enumerate}

\noindent\textbf{Quality control by human review.}
Human reviewers check every derived single-target expression retained in the final dataset. They verify that the rewrite refers to exactly one instance, preserves the key conditions of the original scene, and matches the inherited per-frame boxes. Section~\ref{supp:paired-queries} evaluates the original and rewritten expressions together on the test split.

\paragraph{Zero-target expression construction.}
To train the model to determine whether a long video contains any targets that match a referring expression, we construct 2,013 zero-target expressions. This part of the dataset comprises easy and hard zero-target expressions, which we construct using different procedures. For easy zero-target expressions, we keep the expression unchanged and replace the video that originally matched it with another video. For hard zero-target expressions, we keep the original video clip, first rewrite a multi-target expression to refer to a single target, and then construct a fine-grained, highly confusable counterfactual negative. The construction and quality control procedures are as follows.

\begin{enumerate}
    \item \textbf{Easy zero-target expressions.} We retain the original multi-target expression and its target metadata and pair it with a referring-expression sample from a different video. We first apply a low lexical similarity filter, requiring the maximum TF--IDF cosine similarity for the corresponding group to be at most 0.05. Among the eligible pairs, we prioritize those with higher referring-expression similarity and replace the video associated with the original expression with the video associated with the selected expression. The resulting expression and replacement video are semantically related but do not describe exactly the same content, making the resulting samples moderately challenging.

    \item \textbf{Hard zero-target expressions.} We first construct a single-target expression using the derivation pipeline described above. Given the original expression and frames sampled from the target time interval, Qwen3.8-Max generates three candidate rewrites that do not match the video content. We prioritize replacing the target object or a key contextual object with which it interacts, or changing the appearance, shape, or color of that object to make it clearly distinct from the original. We preserve the main scene, actions, and relationships while introducing a mismatch with the actual video content. Qwen3.8-Flash then selects the best rewrite from the three candidates based on frames sampled at approximately 2 FPS within the target time interval, with a maximum of 32 frames. This design yields challenging counterfactual samples.

    \item \textbf{Quality verification.} We first use a model to check sampled frames for matching targets in both types of zero-target samples, followed by human verification.

    For easy zero-target samples, Qwen3.7-Flash examines up to 64 frames from the receiving video. We sample up to 32 frames at approximately 2 FPS from its existing annotated target time interval, retaining the endpoints, and uniformly sample up to 32 additional frames from outside that interval. The prompt requires that no instance satisfying any target phrase be found for the sample to pass.

    For hard zero-target samples, Qwen3.7-Flash searches the full input clip at approximately 1 FPS using windows of 32 frames with an overlap of two frames between adjacent windows. A sample passes only if no instance satisfying all the conditions of the rewritten expression is found in any sampled window.

    Human reviewers then examine every sample of both types. They manually revise or remove expressions for which they confirm a match.
\end{enumerate}

\begin{table}[!htbp]
    \centering
    \caption{Referring expressions and frame-level boxes by split. Easy zero-target expressions are counted in the split of the replacement clip.}
    \label{tab:supp-split-annotations}
    \small
    \setlength{\tabcolsep}{2.4pt}
    \begin{tabular}{@{}lrrrrrr@{}}
        \toprule
        & \multicolumn{5}{c}{Referring expressions} & \\
        \cmidrule(lr){2-6}
        Split & Single & Multi & Easy zero & Hard zero & All & Boxes \\
        \midrule
        Train      & 1,039 & 3,859 & 838   & 611 & 6,347 & 3,132,546 \\
        Validation & 153   & 566   & 130   & 92  & 941   & 456,276 \\
        Test       & 181   & 977   & 198   & 144 & 1,500 & 834,239 \\
        \midrule
        Total      & 1,373 & 5,402 & 1,166 & 847 & 8,788 & 4,423,061 \\
        \bottomrule
    \end{tabular}
\end{table}

\paragraph{Annotation and screening statistics.}
Table~\ref{tab:supp-annotation-counts} summarizes the recorded stages of annotation and review. These counts aggregate construction records and include items that were later excluded. The final benchmark contains 5,468 directly annotated expressions with matching targets, 1,307 single-target expressions derived from existing annotations, 1,166 easy zero-target expressions, and 847 hard zero-target expressions.

\paragraph{Annotation examples.}
Figure~\ref{fig:supp-annotation-examples-multi} shows how referring expressions identify multiple objects through their appearance, actions, and relationships. Colored boxes link the targets to the corresponding phrases, and yellow bars mark the ground-truth intervals on the full clip timelines.

Figure~\ref{fig:supp-annotation-examples-single-zero} shows annotated single targets alongside two types of zero-target queries. For easy zero-target queries, we show sampled frames from the replacement clip used as input to the model. For hard zero-target queries, the highlighted phrase identifies the changed condition, and the bracketed frames show the conflicting visual evidence.

\subsection{Dataset Statistics and Splits}
\label{supp:statistics}
For the final LV-MOSTG-9K benchmark, Table~\ref{tab:supp-split-sources} lists the video sources for each split, and Table~\ref{tab:supp-split-annotations} reports counts by expression type and the number of frame-level boxes.
Upon acceptance, we will release the LV-MOSTG-9K annotations and official dataset splits, along with the LOMOST code and trained model weights.
The annotations will include source video IDs, clip boundaries, and frame timestamps.

\subsection{Annotation Verification}
\label{supp:annotation-verification}
As described in Sec.~\ref{supp:annotation}, we used four quality control checkpoints during annotation. At checkpoints 1 and 3, checks were performed automatically through a VLM API. At checkpoints 2 and 4, human reviewers checked each item individually.

After annotation, we selected 200 test queries for independent review, stratifying by query type. For queries with matching targets, we further stratified by the fraction of the video duration occupied by the target time interval. To ensure sufficient coverage of zero-target queries, we supplemented the sample drawn in proportion to the query types in the test set with additional zero-target queries. The final sample comprised 121 multi-target, 28 single-target, 21 easy zero-target, and 30 hard zero-target queries, with a slightly higher proportion of hard zero-target queries than in the test set.

Two reviewers independently checked the annotations and agreed on whether any target matched the query for 199/200 queries (99.5\%, Cohen's $\kappa = 0.987$). Following adjudication by a third reviewer and final review, we revised the annotations for 7 queries (agreement on whether to flag an annotation issue was $\kappa = 0.457$). We corrected bounding boxes for 4 queries, extended the target time intervals and annotated the additional frames for 2, and reclassified 1 hard zero-target query as having a matching target. Across the 149 queries originally annotated as having matching targets, the annotations before and after revision had a mean temporal IoU of \textbf{0.9925} and a mean frame-level box-set IoU of \textbf{0.9930} within the original target time intervals.

\section{Training Objectives and Implementation}
\label{supp:training}
\begin{table*}[!tp]
    \centering
    \caption{Training configuration for LOMOST on LV-MOSTG-9K. Checkpoint steps count optimizer updates and identify the weights used for initialization or main-table evaluation.}
    \label{tab:supp-training-config}
    \small
    \setlength{\tabcolsep}{4pt}
    \begin{tabular*}{\textwidth}{@{\extracolsep{\fill}}p{0.16\textwidth}p{0.13\textwidth}p{0.25\textwidth}p{0.16\textwidth}p{0.085\textwidth}p{0.115\textwidth}@{}}
        \toprule
        Stage & Initialization & Training data / input & Optimizer / LR schedule & Effective batch & Checkpoint used \\
        \midrule
        1. Shared image SFT &
        Qwen3-VL-4B-Instruct &
        One pass over 45,000 grounding + 15,000 LLaVA examples\newline
        RGB images, 8,192-token limit &
        AdamW, standard SFT schedule &
        8 examples (1 GPU) & 7,500~steps (best) \\
        2. Mixed video SFT &
        Image SFT (7,500~steps) &
        7,504 BOSTVG matches + 1,251 constructed zero-target examples + 2,501 VidSTG + 1,251 LLaVA\newline
        Full video at 0.5 FPS, 32,768-token limit &
        AdamW, standard SFT schedule &
        8 examples & 1,564~steps (best) \\
        3. LV-MOSTG-9K video SFT &
        Mixed video SFT (1,564~steps) &
        6,344 grounding + 793 LLaVA\newline
        Full clip at 0.5 FPS &
        AdamW, standard SFT schedule &
        8 examples & 600~steps (best) \\
        4. Localization training &
        Frozen VLM (600~steps), new localization weights &
        39,732 planned frame records\newline
        Predicted interval at 2 FPS, nearest-anchor sampling &
        AdamW, constant \(10^{-4}\) &
        32 frames & 1,242~steps (end of run) \\
        \bottomrule
    \end{tabular*}

    \vspace{3pt}
    \begin{minipage}{\textwidth}
        \footnotesize
        Across the three tables, SFT data counts are in examples or labeled exposures, and localization data counts are in frame records. The standard SFT schedule uses 100 warm-up updates to a peak learning rate of \(10^{-5}\), followed by cosine decay to \(10^{-6}\). Weight decay is 0.01 throughout. The shared image SFT checkpoint is both the final checkpoint and the best on validation. LV-MOSTG-9K task-specific video SFT ended at 893 updates.
    \end{minipage}
\end{table*}

\subsection{Loss Definitions and Weights}
\label{supp:losses}
The two objectives in the main paper correspond to task-specific video SFT and localization training, the final two stages of each dataset's training pipeline. All three pipelines begin with shared image SFT. The LV-MOSTG-9K pipeline also includes mixed video SFT before its task-specific video SFT. During localization training, we freeze the resulting VLM and train the Spatial Localization Calibrator and Box Refinement Head with cached VLM outputs and representations.

\paragraph{Task-specific video SFT.}
Let \(\operatorname{CE}(U)\) be the mean next-token cross-entropy over positions in \(U\) for one example. Let \(A\) be the active assistant-answer positions, \(S\subseteq A\) the active \texttt{<SPAT>} positions, and \(E\subseteq A\) the positions encoding the Boolean value of \texttt{confirmed\_empty}. We use
\begin{align}
    \mathcal L_{\mathrm{VLM}} &= \mathcal L_{\mathrm{AR}}+\mathcal L_{\mathrm{token}},\\
    \mathcal L_{\mathrm{AR}} &= \operatorname{CE}(A),\\
    \mathcal L_{\mathrm{token}} &= \operatorname{CE}(S)+0.1\,\operatorname{CE}(E).
\end{align}
Each term is normalized by its own active token count before averaging across examples, and an empty set contributes zero. The answer loss supervises the structured output, including interval and anchor timestamps, coarse box coordinates, and the end-of-sequence token for complete answers. The spatial marker and presence losses give additional weight to positions already covered by the answer loss. LLaVA replay examples use the answer loss alone.

In the formal LV-MOSTG-9K VLM training records, 1,021 examples with spatial-only supervision mask the numeric tokens for the interval and anchor timestamps while retaining box, format, presence, and spatial marker supervision. Another 1,449 zero-target examples supervise a structured empty answer. BOSTVG and VidSTG use the same output schema for expressions with matching targets, with temporal and box tokens supervised.

\paragraph{Structured SFT example.}
The user message pairs sampled video frames with a referring expression, the clip duration, and a frame table:
{\scriptsize
\begin{verbatim}
{"referring_expression": "The red bird on the branch.",
 "duration_sec": 12.0,
 "logical_frames": [
   {"source_frame_idx": 0,   "timestamp_sec": 0.0},
   {"source_frame_idx": 50,  "timestamp_sec": 2.0},
   {"source_frame_idx": 100, "timestamp_sec": 4.0},
   {"source_frame_idx": 150, "timestamp_sec": 6.0}
 ]}
\end{verbatim}
}
The following illustrative target answer contains one matching object at each of the three anchors. Coordinates are normalized and ordered as center \(x\), center \(y\), width, and height. Each anchor can contain multiple matching objects on LV-MOSTG-9K and BOSTVG.
{\scriptsize
\begin{verbatim}
{"proposal": {
  "confirmed_empty": false,
  "time_interval_sec": [1.5, 7.0],
  "anchors": [
    {"frame_id": "frame_0000", "timestamp_sec": 2.0,
     "objects": [{"bbox_cxcywh": [0.41,0.55,0.18,0.39],
                  "spatial_marker": "<SPAT>"}]},
    {"frame_id": "frame_0001", "timestamp_sec": 4.0,
     "objects": [{"bbox_cxcywh": [0.43,0.55,0.18,0.39],
                  "spatial_marker": "<SPAT>"}]},
    {"frame_id": "frame_0002", "timestamp_sec": 6.0,
     "objects": [{"bbox_cxcywh": [0.44,0.56,0.18,0.39],
                  "spatial_marker": "<SPAT>"}]}
  ]
}}
\end{verbatim}
}
For a zero-target expression on LV-MOSTG-9K, the answer is
{\scriptsize
\begin{verbatim}
{"proposal": {"confirmed_empty": true,
              "time_interval_sec": null, "anchors": []}}
\end{verbatim}
}

\paragraph{Localization training.}
The seven loss terms are grouped as follows:
\begin{align}
    \mathcal L_{\mathrm{loc}} &=
    \mathcal L_{\mathrm{box}}+\mathcal L_{\mathrm{cand}}+\mathcal L_{\mathrm{aux}},\\
    \mathcal L_{\mathrm{box}} &= \mathcal L_{\mathrm{reg}}+\mathcal L_{\mathrm{size}},\\
    \mathcal L_{\mathrm{cand}} &= \mathcal L_{\mathrm{map}}+\mathcal L_{\mathrm{center}}
       +0.1\,\mathcal L_{\mathrm{rank}},\\
    \mathcal L_{\mathrm{aux}} &= \mathcal L_{\mathrm{preserve}}+0.1\,\mathcal L_{\mathrm{response}}.
\end{align}
For a matched prediction, let \(\hat{\boldsymbol b}\) be the final box, \(\boldsymbol b_0\) the coarse anchor box, and \(\boldsymbol b^*\) the ground-truth box. Let \(\tilde{\boldsymbol s}\) be the predicted width and height before boundary constraints, and \(\boldsymbol s_0\) and \(\boldsymbol s^*\) the coarse and ground-truth sizes. The box terms are
\begin{align}
    \mathcal L_{\mathrm{reg}} &=
      \tfrac14\|\hat{\boldsymbol b}-\boldsymbol b^*\|_1
      +1-\operatorname{GIoU}(\hat{\boldsymbol b},\boldsymbol b^*),\\
    \mathcal L_{\mathrm{size}} &=
      \tfrac12\|\tilde{\boldsymbol s}-\boldsymbol s^*\|_1.
\end{align}
Let \(p_i\) be the softmax probability of candidate \(i\) over supported cells of the \(15\times15\) grid, using the fused SQCM and FPCM logit \(L_i\). Let \(q_i\) be the corresponding normalized Gaussian target centered at \(\boldsymbol c^*\), with per-axis standard deviation equal to one candidate-grid step. Let \(\bar{\boldsymbol c}=\sum_i p_i\boldsymbol c_i\) be the expected center over the supported cells. The candidate losses are
\begin{align}
    \mathcal L_{\mathrm{map}} &= -\sum_i q_i\log p_i,\\
    \mathcal L_{\mathrm{center}} &=
      \tfrac12\sum_{k\in\{x,y\}}\operatorname{SmoothL1}_{0.1}
      \left(\frac{\bar c_k-c_k^*}{s_{0,k}+10^{-6}}\right),\\
    \mathcal L_{\mathrm{rank}} &=
      \operatorname{softplus}\bigl(1+\max_{i\in N} L_i-L_{i^+}\bigr).
\end{align}
Here \(i^+\) is the nearest supported candidate to the ground-truth center in coarse-box-normalized distance, and \(N\) contains supported candidates outside its \(3\times3\) neighborhood. The map and ranking losses supervise both branches through their fused logits. The center loss uses the full supported grid, while the final prediction reads out a softmax-weighted center in the supported \(3\times3\) neighborhood of the peak.

The auxiliary losses penalize refinements that reduce box overlap and supervise the query response before matching:
\begin{align}
    \mathcal L_{\mathrm{preserve}} &=
      [\operatorname{sg}(\operatorname{IoU}(\boldsymbol b_0,\boldsymbol b^*))
      -\operatorname{IoU}(\hat{\boldsymbol b},\boldsymbol b^*)]_+,\\
    \mathcal L_{\mathrm{response}} &= -\sum_u q^{\mathrm r}_u\log p^{\mathrm r}_u .
\end{align}
Here \(\operatorname{sg}\) stops the gradient and \([z]_+=\max(z,0)\). The response probability \(p^{\mathrm r}_u\) is obtained by softmax over valid cells of the dense query response map. Its Gaussian target \(q^{\mathrm r}_u\) is centered at the ground-truth center, with per-axis standard deviation set to the larger of half the corresponding ground-truth box dimension and one feature-grid step.

\paragraph{Matching and valid supervision.}
LV-MOSTG-9K and BOSTVG use per-frame Hungarian matching to maximize IoU between detached final predictions and the ground-truth box set. VidSTG pairs its single predicted slot directly with the annotated target trajectory at each valid frame. Unmatched slots and frames without valid ground-truth boxes contribute no localization loss. LV-MOSTG-9K zero-target expressions are excluded from this stage. Candidate map, center, and ranking supervision require the ground-truth center to lie within the lattice bounds and at least one supported candidate. Ranking additionally requires a supported negative. Each loss is averaged over its eligible pairs within each frame, then frame losses are averaged.

\subsection{Training Configuration}
\label{supp:training-config}

Tables~\ref{tab:supp-training-config}, \ref{tab:supp-training-bostvg}, and \ref{tab:supp-training-vidstg} list the training stages and checkpoints for LV-MOSTG-9K, BOSTVG, and VidSTG, respectively. Each pipeline begins with the same image SFT checkpoint. The tables number the full training pipelines from this shared stage.

\begin{table*}[!tp]
    \centering
\begin{minipage}[t]{.49\linewidth}
    \vspace{0pt}
    \caption{Training configuration for LOMOST on BOSTVG.}
    \label{tab:supp-training-bostvg}
    \footnotesize
    \begin{tabular}{@{}p{.29\linewidth}@{\hspace{.025\linewidth}}p{.685\linewidth}@{}}
        \toprule
        \multicolumn{2}{@{}l}{\textbf{Stage 1. Shared image SFT}} \\
        \raggedright Initialization & \raggedright Shared with LV-MOSTG-9K (Table~\ref{tab:supp-training-config}). \tabularnewline
        \raggedright Checkpoint used & \raggedright 7,500~steps (best) \tabularnewline
        \midrule
        \multicolumn{2}{@{}l}{\textbf{Stage 2. BOSTVG video SFT}} \\
        \raggedright Initialization & \raggedright Image SFT (7,500~steps) \tabularnewline
        \raggedright Training data / input & \raggedright 7,504 grounding + 938 LLaVA\newline
        Full clip at 0.5 FPS, 32,768-token limit \tabularnewline
        \raggedright Optimizer / LR schedule & \raggedright AdamW, standard SFT schedule, then two cosine continuations \tabularnewline
        \raggedright Effective batch & \raggedright 8 examples \tabularnewline
        \raggedright Checkpoint used & \raggedright 1,806~steps (best) \tabularnewline
        \midrule
        \multicolumn{2}{@{}l}{\textbf{Stage 3. Localization training}} \\
        \raggedright Initialization & \raggedright Frozen VLM (1,806~steps), new localization weights \tabularnewline
        \raggedright Training data / input & \raggedright 39,874 eligible frame records\newline
        Predicted interval at 2 FPS, nearest-anchor sampling \tabularnewline
        \raggedright Optimizer / LR schedule & \raggedright AdamW, constant \(10^{-4}\) \tabularnewline
        \raggedright Effective batch & \raggedright 32 frames \tabularnewline
        \raggedright Checkpoint used & \raggedright 1,247~steps (final) \tabularnewline
        \bottomrule
    \end{tabular}\par
    \vspace{3pt}
    \noindent BOSTVG video SFT ended at 2,049 cumulative optimizer updates, including continuations.
\end{minipage}%
\hfill\begin{minipage}[t]{.49\linewidth}
    \vspace{0pt}
    \caption{Training configuration for LOMOST on VidSTG.}
    \label{tab:supp-training-vidstg}
    \footnotesize
    \begin{tabular}{@{}p{.29\linewidth}@{\hspace{.025\linewidth}}p{.685\linewidth}@{}}
        \toprule
        \multicolumn{2}{@{}l}{\textbf{Stage 1. Shared image SFT}} \\
        \raggedright Initialization & \raggedright Shared with LV-MOSTG-9K (Table~\ref{tab:supp-training-config}). \tabularnewline
        \raggedright Checkpoint used & \raggedright 7,500~steps (best) \tabularnewline
        \midrule
        \multicolumn{2}{@{}l}{\textbf{Stage 2. VidSTG video SFT}} \\
        \raggedright Initialization & \raggedright Image SFT (7,500~steps) \tabularnewline
        \raggedright Training data / input & \raggedright Initial: 10,667 grounding + 1,333 LLaVA. Continuation: 12,000 exposures to fresh grounding and replay data.\newline
        Official segment at 2 FPS, up to 64 frames and 32,768 tokens \tabularnewline
        \raggedright Optimizer / LR schedule & \raggedright AdamW, standard SFT schedule, then constant \(10^{-6}\) \tabularnewline
        \raggedright Effective batch & \raggedright 8 examples \tabularnewline
        \raggedright Checkpoint used & \raggedright 1,750~steps (best) \tabularnewline
        \midrule
        \multicolumn{2}{@{}l}{\textbf{Stage 3. Localization training}} \\
        \raggedright Initialization & \raggedright Frozen VLM (1,750~steps), new localization weights \tabularnewline
        \raggedright Training data / input & \raggedright 59,808 frames used from 10,000 cached training queries\newline
        Same capped 2 FPS frame list, restricted to the predicted interval \tabularnewline
        \raggedright Optimizer / LR schedule & \raggedright AdamW, constant \(10^{-4}\) \tabularnewline
        \raggedright Effective batch & \raggedright 32 frames \tabularnewline
        \raggedright Checkpoint used & \raggedright 1,869~steps (last) \tabularnewline
        \bottomrule
    \end{tabular}\par
    \vspace{3pt}
    \noindent VidSTG video SFT ended at 3,000 cumulative optimizer updates, including the continuation.
\end{minipage}
\end{table*}

\section{Evaluation Protocols and Baseline Adaptation}
\label{supp:evaluation}
\subsection{Dataset-Specific Evaluation}
\label{supp:metrics}
\paragraph{Frame-Level Matching and Spatial Scoring.}
LV-MOSTG-9K provides box annotations at the native video frame rate, allowing evaluation on a chosen subset of frames or on all native frames. In this paper, we evaluate all methods on a common 2~FPS grid. For each scoring frame $f$ of query $q$, let $A_{qf}$ and $B_{qf}$ be the ground-truth and predicted box sets. Each set is empty outside its respective time interval. We apply Hungarian matching independently in each frame to maximize the sum of IoUs between matched boxes. Let $M_{qf}$ denote the resulting set of matched pairs. The intersection and union areas for frame $f$ are
\begin{align}
    I_{qf}
      &= \sum_{(a,b)\in M_{qf}}\operatorname{area}(a\cap b),\\
    U_{qf}
      &= \sum_{a\in A_{qf}}\operatorname{area}(a)
        + \sum_{b\in B_{qf}}\operatorname{area}(b)-I_{qf}.
\end{align}
The union term $U_{qf}$ includes the areas of unmatched boxes. If both box sets are empty, the frame contributes zero to both area sums.

\paragraph{LV-MOSTG-9K Grounding Metrics.}
Let $\mathcal{Q}_+$ denote the 1,158 test queries with matching targets. For each query $q$, let $G_q$ and $P_q$ be the ground-truth and predicted time intervals, respectively. Both intervals are half-open.

\paragraph{1. tIoU and m-tIoU.}
Temporal IoU measures the overlap between the predicted and ground-truth intervals.
\begin{align}
    \operatorname{tIoU}(q)
      &= \frac{|G_q\cap P_q|}{|G_q\cup P_q|},\\
    \operatorname{m\text{-}tIoU}
      &= \frac{1}{|\mathcal{Q}_+|}
         \sum_{q\in\mathcal{Q}_+}\operatorname{tIoU}(q).
\end{align}
In the first equation, $|\cdot|$ denotes interval duration. The query-level tIoU is zero if the predicted interval is missing, invalid, or disjoint from the ground-truth interval. We compute m-tIoU as the arithmetic mean of query-level tIoU.

\paragraph{2. vIoU and m-vIoU.}
For each query, we sum intersection and union areas over the scoring grid before taking their ratio. A zero denominator yields a score of zero. We obtain m-vIoU by averaging the query-level scores.
\begin{align}
    \operatorname{vIoU}(q)
      &= \frac{\sum_f I_{qf}}{\sum_f U_{qf}},\\
    \operatorname{m\text{-}vIoU}
      &= \frac{1}{|\mathcal{Q}_+|}
         \sum_{q\in\mathcal{Q}_+}\operatorname{vIoU}(q).
\end{align}

\paragraph{3. vIoU@$R$.}
The LOMOST scorer reports the fraction of queries whose vIoU strictly exceeds $R$, for $R\in\{0.3,0.5\}$.
\begin{equation}
    \operatorname{vIoU@}R =
    \frac{1}{|\mathcal{Q}_+|}\sum_{q\in\mathcal{Q}_+}
    \mathbf{1}\!\left[\operatorname{vIoU}(q)>R\right].
\end{equation}

\paragraph{4. vIoU-Int.\ All.}
Let $H_q=\{f:t_f\in G_q\cap P_q\}$ denote the scoring frames in the intersection of the predicted and ground-truth intervals, where $t_f$ is the frame timestamp. For each query, we average frame-level spatial IoU over $H_q$.
\begin{equation}
    \operatorname{vIoU\text{-}Int}(q)
      = \frac{1}{|H_q|}\sum_{f\in H_q}
        \begin{cases}
            I_{qf}/U_{qf}, & U_{qf}>0,\\
            0, & U_{qf}=0.
        \end{cases}
\end{equation}
Frames with two empty box sets count as zero in the frame average. If $H_q$ is empty or spatial prediction fails, the query-level score is zero. The suffix \emph{All} denotes averaging over all 1,158 queries with matching targets.
\begin{equation}
    \operatorname{vIoU\text{-}Int.\ All}
      = \frac{1}{|\mathcal{Q}_+|}\sum_{q\in\mathcal{Q}_+}
        \operatorname{vIoU\text{-}Int}(q).
\end{equation}
Excluding frames where both box sets are empty preserves the vIoU-Int.\ All ranking of all 12 methods in Table~2 of the main paper and changes scores by at most 0.42 percentage points.

We report the LV-MOSTG-9K scores as percentages. A query with a valid predicted interval and a failed spatial prediction retains its tIoU, with spatial scores set to zero.

\paragraph{Zero-Target Recognition.}
We evaluate zero-target recognition on all 1,500 test queries and treat the 342 queries with no matching targets as the positive class. We count a prediction as zero-target when the evaluation record is valid, has a null predicted interval, and contains no predicted boxes on any scoring frame. Sec.~\ref{supp:losses} gives an example of LOMOST's structured empty output. A missing or unparseable response fails to register a zero-target prediction and counts as a false negative on a zero-target query. We report precision, recall, and F1.

\paragraph{BOSTVG.}
We evaluate LOMOST with the scorer from the official BOSTVG repository. Citation numbers in this supplement refer to the reference list in the main paper. The baseline scores in our BOSTVG table come from OmniSTVG~\suppmaincite{yao2026omnistvg}, which uses the same official evaluation protocol.

\paragraph{VidSTG.}
On the official VidSTG test split, all methods use the same input frames, sampled at 2~FPS and uniformly subsampled if the count exceeds 128. When reconstructing a trajectory, we linearly interpolate between adjacent valid box predictions using the original frame timestamps. We score all methods at the same 2~FPS timestamps with no cap on the number of scoring frames.

\subsection{Task-Specific Baselines}
\label{supp:baselines}
For LV-MOSTG-9K, we initialize TubeDETR, STCAT, and TA-STVG from the official VidSTG checkpoints released by their respective authors and train the adapted models on the LV-MOSTG-9K training split. Section~\ref{supp:training-config} details the initialization and training data for each stage of LOMOST.
Table~\ref{tab:supp-lv-baseline-adaptation} summarizes the shared and model-specific changes made to adapt the baselines to LV-MOSTG-9K.

\begin{table*}[!tp]
    \centering
    \caption{Adaptation of task-specific baselines to LV-MOSTG-9K. The shared settings apply to the three methods with results on TEST1500. The CG-STVG entry describes the configuration used for validation diagnostics and efficiency measurements.}
    \label{tab:supp-lv-baseline-adaptation}
    \small
    \setlength{\tabcolsep}{4pt}
    \renewcommand{\arraystretch}{1.10}
    \begin{tabular*}{\textwidth}{@{\extracolsep{\fill}}p{0.17\textwidth}p{0.43\textwidth}p{0.34\textwidth}@{}}
        \toprule
        \multicolumn{3}{@{}l@{}}{\textbf{Shared adaptation for TubeDETR, STCAT, and TA-STVG}} \\
        \midrule
        Long-video input &
        \multicolumn{2}{p{0.79\textwidth}@{}}{Sample the full clip at 2~FPS and process it in 25-second windows with a 12.5-second stride.} \\
        Multi-target output &
        \multicolumn{2}{p{0.79\textwidth}@{}}{Use 12 candidate slots per frame. Each window predicts one time interval shared by its slots.} \\
        Training supervision &
        \multicolumn{2}{p{0.79\textwidth}@{}}{Match predicted slots to ground-truth boxes independently in each frame using Hungarian matching. Zero-target queries supervise only window-level rejection. Background windows verified to have no match also supervise empty box sets on all valid frames.} \\
        Zero-target prediction &
        \multicolumn{2}{p{0.79\textwidth}@{}}{Return a null interval and no boxes when no window passes the confidence threshold for a match.} \\
        \midrule
        \multicolumn{3}{@{}l@{}}{\textbf{Model-specific adaptations}} \\
        \midrule
        Method & Multi-target extension & Temporal prediction \\
        \midrule
        TubeDETR &
        Retain the fast and slow video encoding streams. Add a learnable offset to the original query for each slot and predict boxes and objectness per slot. &
        Average the decoded slot features at each frame and pass them to the original start and end prediction heads. \\
        STCAT &
        Add learnable four-dimensional slot offsets to the reference box from the template generator. Refine the slot-specific boxes with the shared spatial decoder. &
        Run the original temporal decoder once per window. \\
        TA-STVG &
        Use the full spatio-temporal decoder for the first slot and extend the spatial queries for the remaining slots. &
        Use the first slot's temporal features for the shared interval. During inference, refine the queries in a second pass using frames selected by first-pass actionness scores. \\
        \midrule
        CG-STVG\newline (engineering adaptation) &
        Extract appearance and motion features within each slot's predicted box. Use quality scores to select context for the next decoder layer. &
        Diagnose temporal selection and window merging on the validation split. No formal TEST1500 accuracy is reported. \\
        \bottomrule
    \end{tabular*}
\end{table*}

\paragraph{CG-STVG on LV-MOSTG-9K.}
Our CG-STVG adaptation showed overlapping predictions across object slots and missed intervals after window merging on the validation set. We therefore omit it from the LV-MOSTG-9K test accuracy table.

\paragraph{VidSTG.}
We re-evaluate TubeDETR, STCAT, CG-STVG, and TA-STVG using their official checkpoints. All four baselines receive the same frames sampled at a target rate of 2~FPS, with a cap of 128 input frames. We score spatial predictions on the shared 2~FPS frame set without a frame cap. For TA-STVG, we prepend the subject field from the authors' test metadata to the referring expression.

\paragraph{BOSTVG.}
We take the results for adapted TubeDETR, STCAT, and CG-STVG, as well as OmniTube, from OmniSTVG~\suppmaincite{yao2026omnistvg}. The BOSTVG baseline is the authors' simplified OmniTube variant with the text-guided object query generation module and spatio-temporal attention blocks removed. Its results come from Table~3(d) of an earlier public version of OmniSTVG.

\subsection{Vision-Language and API Baselines}
\label{supp:vlm-baselines}
For the evaluation of general-purpose VLMs on LV-MOSTG TEST1500, we run GLM-4.6V-Flash (9B), InternVideo3-8B, Qwen3.5-4B, and LLaVA-OneVision-2-8B locally and access Qwen3.5-27B, Qwen3.5-122B-A10B, Gemini 3.5 Flash-Lite, and GPT-5.6 Luna through cloud APIs. All eight general-purpose VLM baselines in Table~2 of the main paper receive the text query and video frames sampled globally at 0.5~FPS, matching the global sampling rate of LOMOST. The model predicts the target time interval and bounding boxes on keyframes. We match predicted boxes across adjacent keyframes using Hungarian matching on box IoU, linearly interpolate pairs with positive IoU, and copy unmatched boxes from the nearer keyframe. We evaluate the reconstructed boxes on a common 2~FPS time grid.
For the task-specific fine-tuned Qwen3-VL-4B baseline, Sec.~\ref{supp:spatial-stability} (Table~\ref{tab:supp-sft-linear}) compares SFT + Box Copy from the ablation study in Sec.~5.3 of the main paper with SFT + Linear, both using the same 0.5~FPS global sampling rate as the VLM baselines above.

\begin{figure}[!tp]
    \centering
    \includegraphics[width=\columnwidth]{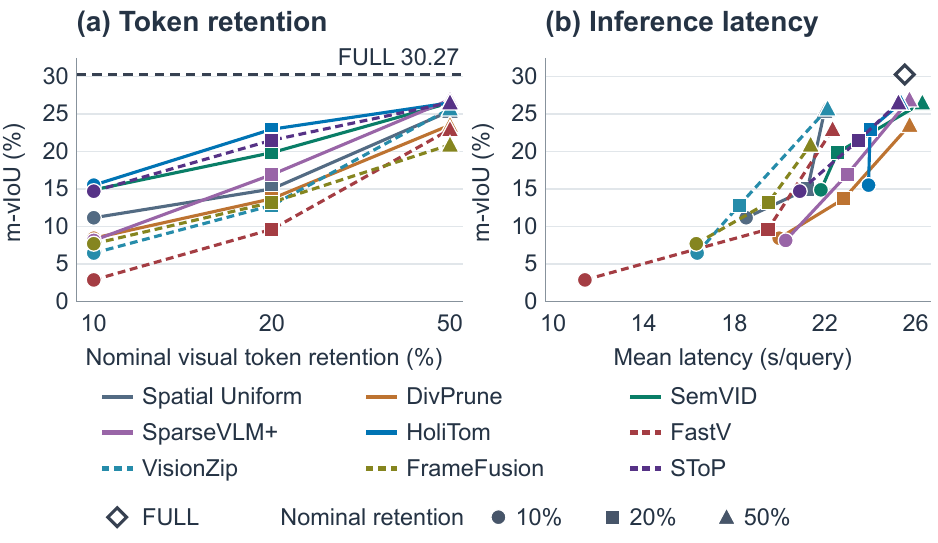}
    \caption{Visual token compression on LV-MOSTG-9K. (a) m-vIoU on TEST1500 at three equally spaced nominal retention settings. (b) the same scores against mean latency measured on 32 queries. Lines connect the three settings for each method. The dashed line and open diamond indicate uncompressed LOMOST (FULL).}
    \label{fig:supp-compression-tradeoff}
\end{figure}

\section{Efficiency and Visual Token Compression}
\label{supp:efficiency-compression}
\subsection{Efficiency Protocol and Runtime Breakdown}
\label{supp:efficiency}
\paragraph{Evaluation setup.}
Tables~6 and~7 of the main paper both use a single NVIDIA A800 with a batch size of 1, and we time each query individually after warm-up. For efficiency evaluation, we select 32 representative queries by stratifying on video duration and including zero-target, single-target, and multi-target samples to reflect the overall distribution of the dataset. We report the mean runtime per query, P95 latency, and the maximum peak allocated GPU tensor memory across the 32 queries.

\paragraph{Runtime breakdown.}
LOMOST in main-paper Table~6 (45.35\,s) and FULL in main-paper Table~7 (25.52\,s) use the same weights, but the former includes decoding video files and extracting frames, whereas the latter includes reading and decompressing a prebuilt RGB frame cache, with the largest difference between the two runs occurring in global frame loading and preprocessing. For these two runs, respectively, frame loading and preprocessing took 24.66/3.23 s per query, VLM generation took 15.84/19.17 s, and the remaining steps took 4.85/3.12 s.

For all methods in Table~6 of the main paper, end-to-end latency includes video decoding, frame sampling, and preprocessing. Compared with TA-STVG, the strongest task-specific baseline by m-vIoU on LV-MOSTG-9K, LOMOST takes only 3.02\,s longer per query on average (45.35 vs.\ 42.33\,s) and raises m-vIoU from 23.64\% to 30.27\%, a gain of 6.63 percentage points. This corresponds to a 7.1\% increase in mean latency and a 28.0\% relative gain in m-vIoU.

\begin{figure}[!htbp]
\centering
\includegraphics[width=\columnwidth]{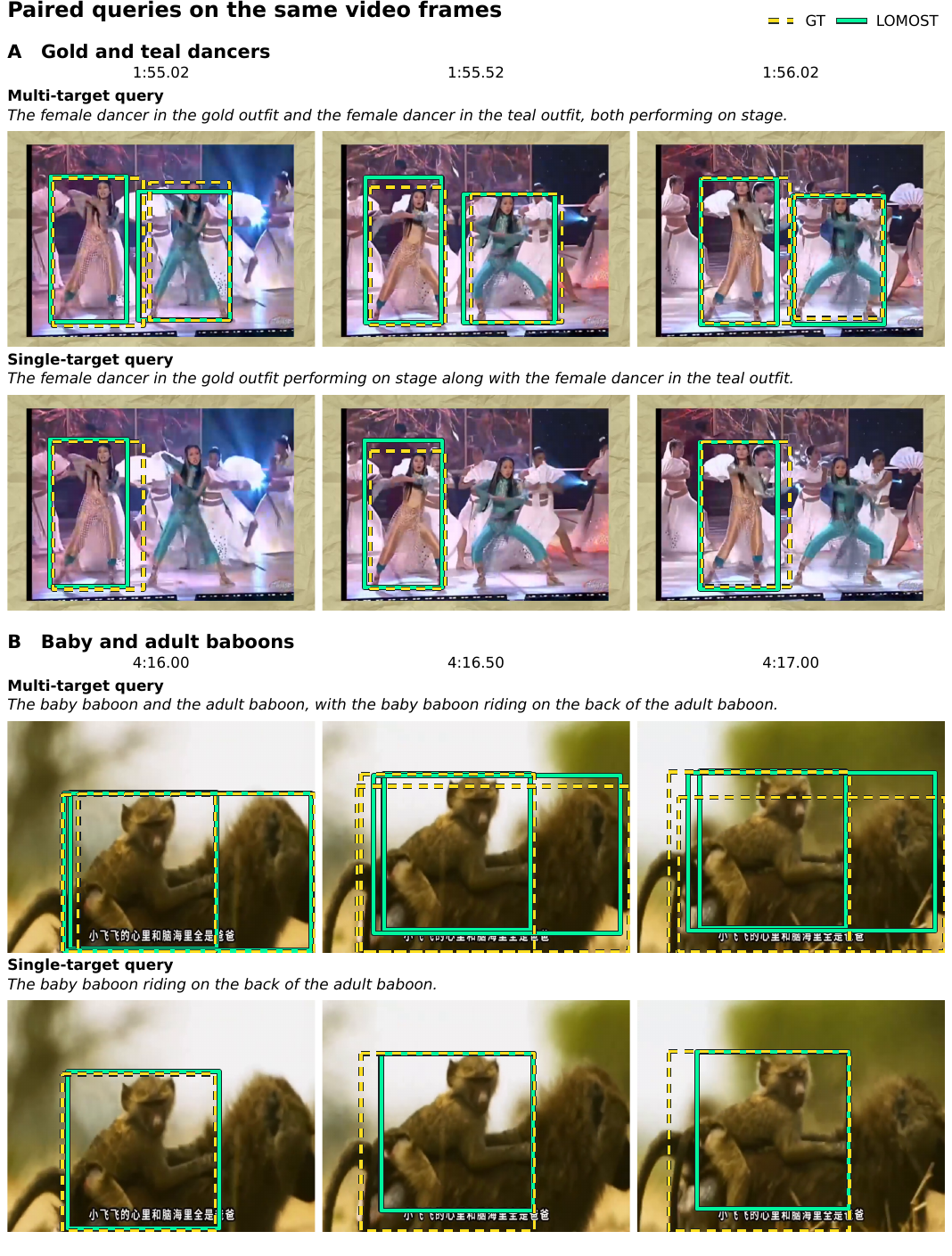}
\caption{Paired queries on identical frames. The multi-target query refers to both dancers in A and both baboons in B. The single-target query retains the gold-clad dancer or the baby baboon and uses the other instance as context. Dashed yellow boxes show ground truth and solid green boxes show LOMOST predictions. We selected examples from annotations before inspecting predictions and show the three eligible frames nearest each selected timestamp. Both dancer rows use the same crop.}
\label{fig:paired-queries}
\end{figure}

\subsection{Compression Setup and Additional Results}
\label{supp:compression}
We apply visual token compression only in the global VLM, and the localizer continues to use the original reference-frame and local visual features.
We evaluate DivPrune, SemVID, SparseVLM+, HoliTom, FastV, VisionZip, FrameFusion, and SToP, along with the Spatial Uniform baseline, with their adaptations to LOMOST described in Table~\ref{tab:supp-compression-adaptation}.

To examine which visual token compression strategies are best suited to GL-STVG, we select eight representative compression methods developed for general-purpose vision-language models and include Spatial Uniform as a baseline. We adapt all nine strategies to LOMOST and compare their grounding accuracy and inference latency on LV-MOSTG-9K. Due to space constraints, Table~7 in the main paper reports results for five of these methods. Table~\ref{tab:supp-compression-results} and Figure~\ref{fig:supp-compression-tradeoff} in the supplementary material add results for FastV, VisionZip, FrameFusion, and SToP, providing a comparison of grounding accuracy and inference latency across all nine methods on LV-MOSTG-9K.

\textbf{The target interval may occupy only a small fraction of a long video, so how a limited token budget is distributed across segments affects temporal grounding.} SemVID allocates tokens across segments and frames based on the referring expression and inter-frame variation. It achieves the highest overall m-tIoU at both 10\% and 50\% retention. At 10\% retention, SemVID achieves an m-tIoU of 24.66\% on target intervals lasting only 2--4 seconds, compared with 15.50\% for Spatial Uniform. One possible explanation is that when most of the footage is irrelevant to the query, query-guided allocation helps preserve cues that mark when a target appears, an action changes, or an interaction ends, allowing the model to locate temporal boundaries more precisely.

\textbf{For multi-target queries, the model must also preserve cues about the location and interactions of each instance.} HoliTom reduces redundant observations through spatio-temporal merging and achieves the highest overall m-vIoU at both 10\% and 20\% retention. On multi-target queries at 20\% retention, HoliTom achieves an m-vIoU of 22.36\%, compared with 14.45\% for Spatial Uniform. Adjacent frames in long videos often share similar backgrounds. Merging these repeated observations may allow the model to retain more visual information useful for grounding within a limited budget. Figure~\ref{fig:supp-compression-tradeoff} also illustrates how the trade-off between accuracy and latency varies with the placement of compression and the token budget. FastV reduces the visual tokens to the target budget immediately after the second language-model block. At 10\% retention, it has the lowest latency, but its m-vIoU falls to 2.89\%. One possible explanation is that aggressive early compression discards cues about briefly visible targets, making it difficult for later layers to predict the correct time interval and anchor boxes. At the more generous 50\% retention rate, SparseVLM+, which prunes tokens progressively at three layers, achieves the highest m-vIoU (26.92\%). These results suggest that compression strategies for GL-STVG should consider \textbf{when to prune, which video segments are relevant to the query, and whether the retained visual cues cover multiple target instances}.

\begin{figure*}[!tp]
    \centering
    \includegraphics[width=\textwidth,height=0.91\textheight,keepaspectratio]{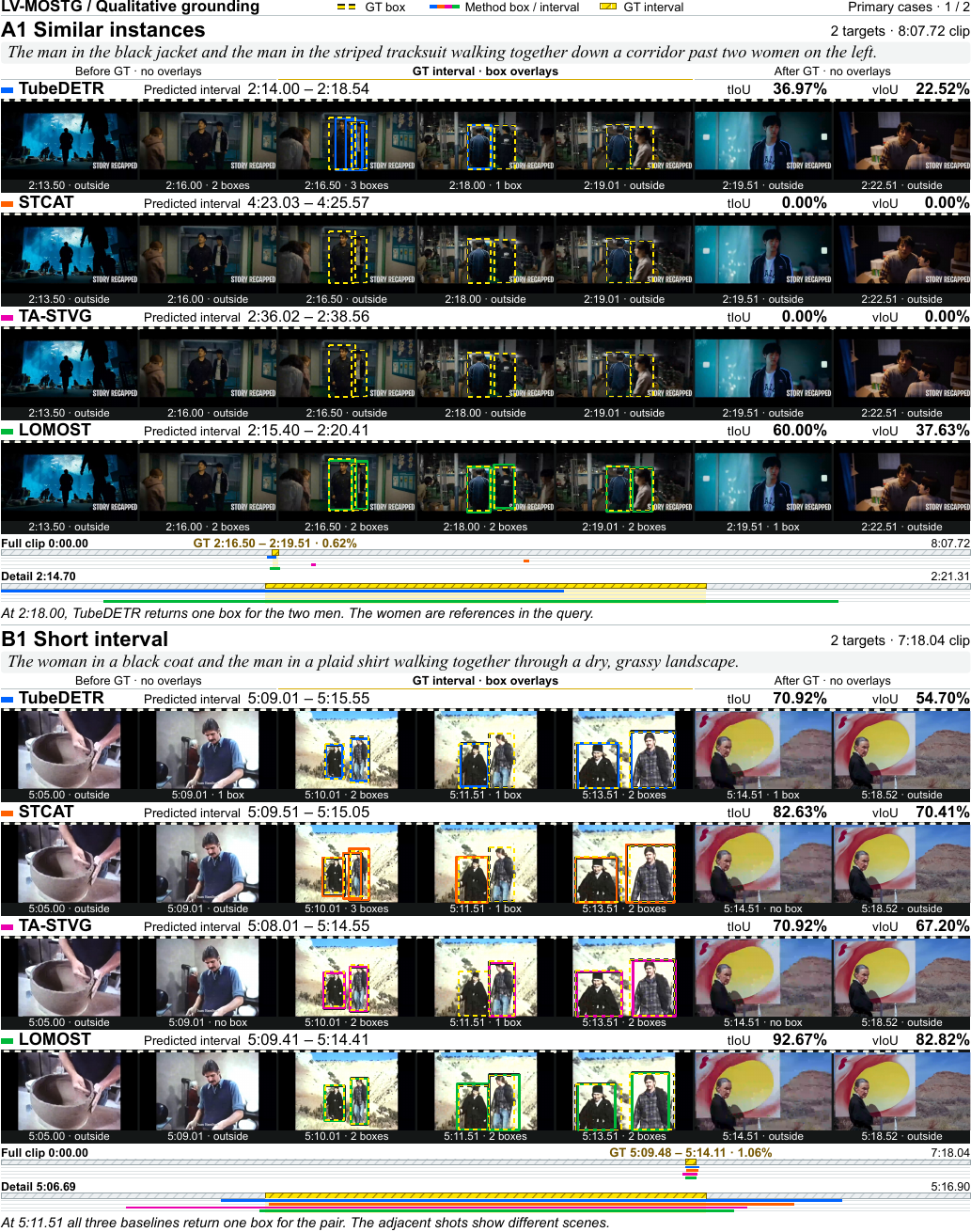}
    \caption{LV-MOSTG-9K predictions for two multi-target queries. The first requires distinguishing similar instances, and the second requires localizing a short target time interval.}
    \label{fig:supp-qualitative-similar-short}
\end{figure*}

\begin{figure*}[!tp]
    \centering
    \includegraphics[width=\textwidth,height=0.91\textheight,keepaspectratio]{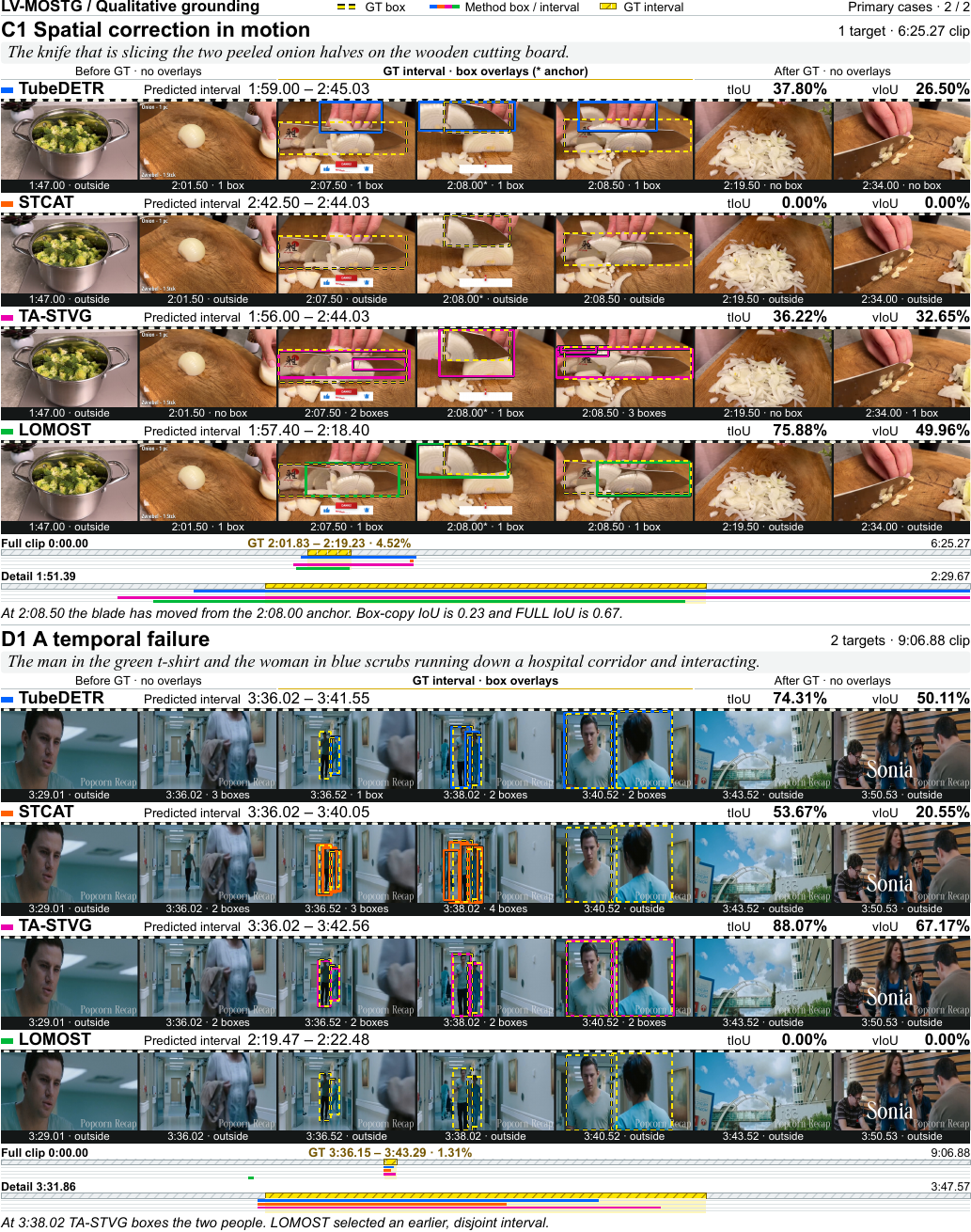}
    \caption{Spatial calibration for a moving target (top) and a temporal localization failure (bottom). In the latter case, LOMOST predicts an earlier interval disjoint from the ground truth.}
    \label{fig:supp-qualitative-motion-failure}
\end{figure*}

\begin{table*}[!tp]
    \centering
    \caption{Integration of nine visual token compression methods into LOMOST. LM layer indices start at zero. Pre-LM compression takes place after the visual merger and before the first language-model block.}
    \label{tab:supp-compression-adaptation}
    \small
    \setlength{\tabcolsep}{4pt}
    \renewcommand{\arraystretch}{1.10}
    \begin{tabular*}{\textwidth}{@{\extracolsep{\fill}}p{0.16\textwidth}p{0.25\textwidth}p{0.55\textwidth}@{}}
        \toprule
        Method & Integration point & Adaptation to LOMOST \\
        \midrule
        Spatial Uniform &
        Pre-LM &
        We keep LOMOST's global frame sampling and distribute the target token budget across the sampled frames. \\
        DivPrune~\suppmaincite{alvar2025divprune} &
        Pre-LM &
        We run the selector on Qwen3-VL's projected visual features and retain tokens according to the target budget. \\
        SemVID~\suppmaincite{li2026keeping} &
        Pre-LM &
        We pass the referring expression embeddings and global visual features to SemVID to allocate segment-level and frame-level budgets. \\
        SparseVLM+~\suppmaincite{zhang2024sparsevlm} &
        LM layers 2, 6, and 15 &
        We rescale the three stage budgets to the target token count, retain Qwen3-VL's mRoPE, and disable token recycling and merging. \\
        HoliTom~\suppmaincite{shao2505holitom} &
        Pre-LM, then LM layer 18 &
        For each target budget, we preset a retention rate before the LM and a drop rate at the internal compression stage. \\
        FastV~\suppmaincite{chen2024image} &
        After the second LM block &
        We reuse the same retained token indices during generation and hidden-state recomputation. \\
        VisionZip~\suppmaincite{yang2025visionzip} &
        Pre-LM &
        We use attention scores from the final layer of Qwen3-VL's visual encoder and merge DeepStack features using the same token assignments. \\
        FrameFusion~\suppmaincite{fu2025framefusion} &
        Inside the LM at multiple points as needed &
        We cap the number of tokens removed at each step based on the current and target token counts. \\
        SToP~\suppmaincite{kim2026sink} &
        Pre-LM, then LM layer 18 &
        We set the pre-LM budget based on available tokens and compute the internal drop rate from the number of tokens retained by the first stage and the target count. \\
        \bottomrule
    \end{tabular*}
\end{table*}

\begin{table}[!tp]
    \centering
    \caption{Visual token compression results grouped by nominal retention. Actual retention is the ratio of retained to original global visual tokens, pooled over TEST1500. Accuracy is reported on TEST1500, with the FULL accuracy taken from Table~2 of the main paper. Efficiency is measured on 32 queries on one NVIDIA A800 at batch size 1. Peak denotes the maximum allocated GPU memory across queries. Within each nominal retention group, bold and underlined entries denote the best and second-best results, respectively, ranked before rounding (higher accuracy and lower latency and memory are better).}
    \label{tab:supp-compression-results}
    \scriptsize
    \setlength{\tabcolsep}{1pt}
    \begin{tabular*}{\columnwidth}{@{\extracolsep{\fill}}lrrrrrrrr@{}}
        \toprule
        & & \multicolumn{4}{c}{Accuracy (\%)} & \multicolumn{3}{c}{Efficiency} \\
        \cmidrule(lr){3-6}\cmidrule(l){7-9}
        Method & \shortstack{Actual\\(\%)} & m-tIoU & m-vIoU & \shortstack{vIoU-Int.\\All} & \shortstack{Zero-\\target F1} & \shortstack{Mean\\(s)} & \shortstack{P95\\(s)} & \shortstack{Peak\\(GiB)} \\
        \midrule
        FULL (100\%) & 100.00 & 49.34 & 30.27 & 48.16 & 92.51 & 25.52 & 50.66 & 19.41 \\
        \midrule
        \multicolumn{9}{@{}l}{\textit{10\% nominal retention}} \\
        Spatial Uniform & 10.00 & 29.59 & 11.18 & 19.84 & 83.66 & 18.52 & 44.49 & 15.37 \\
        DivPrune & 10.00 & 24.97 & 8.45 & 17.26 & 89.99 & 19.98 & 39.93 & 23.85 \\
        SemVID & 10.00 & \textbf{37.61} & \underline{14.89} & 23.61 & 89.77 & 21.82 & 39.23 & 15.37 \\
        SparseVLM+ & 10.00 & 21.54 & 8.15 & 15.21 & 71.30 & 20.27 & 48.26 & 42.30 \\
        HoliTom & 10.20 & \underline{37.52} & \textbf{15.53} & \textbf{25.64} & \underline{90.41} & 23.92 & 55.94 & \textbf{15.36} \\
        FastV & 10.00 & 8.22 & 2.89 & 6.79 & 55.09 & \textbf{11.40} & \textbf{29.26} & \underline{15.37} \\
        VisionZip & 10.00 & 19.28 & 6.47 & 11.81 & 74.47 & 16.36 & 40.25 & 15.37 \\
        FrameFusion & 10.00 & 20.22 & 7.71 & 15.84 & 76.50 & \underline{16.32} & \underline{31.49} & 21.49 \\
        SToP & 10.00 & 36.57 & 14.72 & \underline{23.79} & \textbf{91.54} & 20.88 & 38.15 & 15.56 \\
        \midrule
        \multicolumn{9}{@{}l}{\textit{20\% nominal retention}} \\
        Spatial Uniform & 20.00 & 36.76 & 15.01 & 24.67 & 89.17 & 21.23 & 43.58 & 15.42 \\
        DivPrune & 20.00 & 34.01 & 13.71 & 24.63 & \underline{91.92} & 22.83 & 48.38 & 23.89 \\
        SemVID & 20.00 & 42.48 & 19.86 & 31.47 & 91.68 & 22.55 & 42.21 & 15.42 \\
        SparseVLM+ & 20.00 & 35.65 & 16.93 & 28.98 & 84.04 & 22.97 & 40.01 & 42.30 \\
        HoliTom & 20.15 & \textbf{44.98} & \textbf{22.98} & \textbf{36.34} & \textbf{92.12} & 23.99 & 43.11 & \textbf{15.36} \\
        FastV & 20.00 & 21.19 & 9.64 & 18.96 & 70.58 & \underline{19.49} & 40.01 & \underline{15.41} \\
        VisionZip & 20.00 & 32.43 & 12.82 & 21.84 & 86.76 & \textbf{18.21} & \textbf{35.98} & 15.59 \\
        FrameFusion & 20.00 & 30.67 & 13.23 & 24.27 & 85.79 & 19.51 & \underline{37.64} & 19.79 \\
        SToP & 20.00 & \underline{43.89} & \underline{21.51} & \underline{34.14} & 91.74 & 23.48 & 45.94 & 15.62 \\
        \midrule
        \multicolumn{9}{@{}l}{\textit{50\% nominal retention}} \\
        Spatial Uniform & 50.00 & \underline{46.79} & 25.40 & 39.69 & 91.23 & \underline{21.98} & \underline{42.32} & \textbf{15.86} \\
        DivPrune & 50.00 & 45.25 & 23.53 & 37.99 & \underline{92.95} & 25.73 & 52.21 & 23.89 \\
        SemVID & 50.00 & \textbf{47.29} & 26.50 & 41.49 & 92.40 & 26.30 & 51.35 & \textbf{15.86} \\
        SparseVLM+ & 50.00 & 46.61 & \textbf{26.92} & \textbf{43.91} & 91.63 & 25.73 & 49.62 & 42.94 \\
        HoliTom & 49.83 & 46.67 & 26.50 & \underline{41.85} & 92.19 & 25.36 & 45.00 & \underline{15.98} \\
        FastV & 50.00 & 40.38 & 22.95 & 39.41 & 91.09 & 22.33 & 47.00 & 15.98 \\
        VisionZip & 50.00 & 44.59 & 25.71 & 41.02 & 91.88 & 22.11 & 44.98 & 16.02 \\
        FrameFusion & 50.00 & 42.48 & 20.90 & 33.79 & 92.70 & \textbf{21.36} & \textbf{40.12} & 22.41 \\
        SToP & 50.00 & 46.02 & \underline{26.53} & 41.80 & \textbf{93.22} & 25.24 & 46.83 & 16.93 \\
        \bottomrule
    \end{tabular*}
\end{table}

\section{Additional Analysis}
\label{supp:additional-analysis}
\subsection{Grounding under Paired Single- and Multi-Target Queries}
\label{supp:paired-queries}

We pair 119 of the 181 TEST1500 single-target queries with their original multi-target expressions (112 clips, 95 source videos). Each pair shares the clip, interval, and boxes of the retained instance. The single-target query mentions the other instances as context.

\paragraph{Evaluation protocol.}
We reuse the final TEST1500 predictions for all 238 queries. On the 2~FPS grid used in our experiments, we score frames within the shared ground-truth interval where both the retained instance and at least one context instance are visible. All pairs contribute eligible frames (2,314 in total). Each query's prediction is empty outside its predicted interval. A predicted set is exact if all predicted and visible ground-truth boxes match one-to-one at IoU $\geq 0.5$, with no extra or missed boxes. Paired exact-set accuracy requires both queries to be exact on the same frame. For context leakage on the single-target query, we reserve the best qualifying prediction for the retained instance, match the remaining predictions to visible context instances, and count the fraction of visible context instances matched. We compute each rate within a pair and average equally over pairs.

\begin{table}[!htbp]
\centering
\caption{Grounding accuracy on paired TEST1500 queries (\%). Single, Multi, and Paired report exact-set accuracy. Context is the rate of localizing a context-only instance under the single-target query.}
\label{tab:paired-set-grounding}
\small
\setlength{\tabcolsep}{3pt}
\begin{tabular*}{\linewidth}{@{\extracolsep{\fill}}lrrrr@{}}
\toprule
Method & Single $\uparrow$ & Multi $\uparrow$ & Paired $\uparrow$ & Context $\downarrow$ \\
\midrule
TA-STVG & 28.20 & 20.95 & 9.49 & 12.64 \\
SFT + Box Copy & 45.01 & 28.93 & 23.75 & \textbf{2.25} \\
LOMOST & \textbf{48.98} & \textbf{32.24} & \textbf{25.81} & 2.89 \\
\bottomrule
\end{tabular*}
\par

\end{table}

\paragraph{Results.}
LOMOST achieves 25.81\% paired exact-set accuracy, compared with 9.49\% for TA-STVG (Table~\ref{tab:paired-set-grounding}). The difference is 16.32 percentage points (95\% CI $[11.25,21.62]$). LOMOST also localizes context-only instances less often (2.89\% vs.\ 12.64\%). When scoring target counts alone, LOMOST and SFT + Box Copy both reach 58.00\% paired accuracy, compared with 12.65\% for TA-STVG. Spatial refinement preserves the predicted counts and yields an observed gain of 2.05 points in paired exact-set accuracy over box copying (95\% CI $[-0.59,4.44]$). Both intervals use 20,000 paired bootstrap resamples of source videos. Figure~\ref{fig:paired-queries} illustrates how predictions change when a participant serves as context.

\subsection{Easy and Hard Zero-Target Queries}
\label{supp:zero-split}
All 12 models reject cross-video negatives more reliably than same-video counterfactuals (Fig.~\ref{fig:zero-split}). This ordering is consistent with the construction of hard negatives, which preserve most of the scene and description while changing one condition, such as an object's color, action, or state. As described in Sec.~\ref{supp:annotation}, we use the Qwen3.7-Flash VLM API to check sampled windows across the full clip for matches to each hard negative, followed by human verification of every sample. LOMOST correctly rejects 196/198 easy negatives (98.99\%) and 119/144 hard negatives (82.64\%), with 25 of its 27 missed rejections occurring on hard queries. It falsely rejects 24/1,158 target-present queries (2.07\%).

The false rejection rate helps interpret the high easy-negative scores. Qwen3.5-4B nearly matches LOMOST on easy negatives (98.48\%) but rejects 26.17\% of target-present queries, the highest rate among the three Qwen3.5 variants. Positive-query grounding scores also reveal little about this behavior in the two API models: Gemini 3.5 Flash-Lite and GPT-5.6 Luna attain similar m-tIoU (45.84\% and 47.33\%) and zero-target precision (both approximately 89\%), yet correctly reject 241 and 103 of the same 342 negatives, respectively (70.47\% vs.\ 30.12\% recall). All rates follow the TEST1500 scoring rules in Sec.~C.1 and retain missing and invalid outputs in the denominators.

\subsection{Qualitative Results}
\label{supp:qualitative-results}
Figures~\ref{fig:supp-qualitative-similar-short} and~\ref{fig:supp-qualitative-motion-failure} show a visual comparison of predictions from LOMOST and three adapted spatio-temporal grounding baselines on four representative test queries from LV-MOSTG-9K.

\begin{figure*}[!tp]
    \begin{minipage}[t]{.49\textwidth}\vspace{0pt}
\centering
    \includegraphics[width=\linewidth]{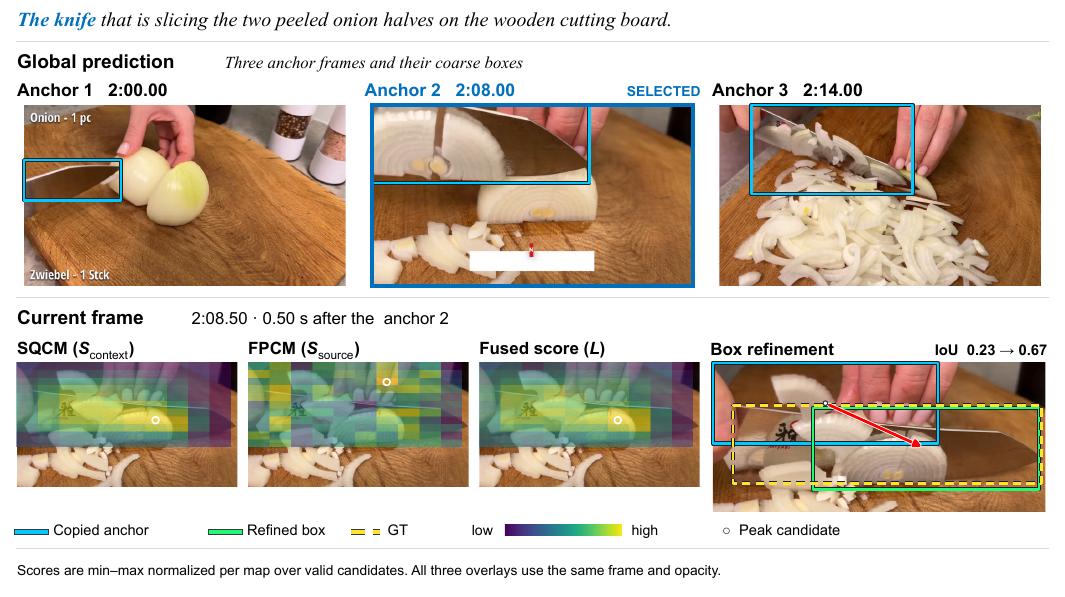}
    \caption{Spatial calibration for the moving knife in case C1. The top row shows the three predicted anchor boxes, with the selected anchor at 2:08.00 highlighted. The bottom row shows the SQCM, FPCM, and fused candidate scores on the frame at 2:08.50, followed by the copied anchor box, refined prediction, and ground truth. The arrow shows the shift in box center, and the refined box improves frame-level IoU from 0.23 to 0.67.}
    \label{fig:supp-spatial-calibration}
\end{minipage}\hfill
\begin{minipage}[t]{.49\textwidth}\vspace{0pt}
\centering
\includegraphics[width=\linewidth]{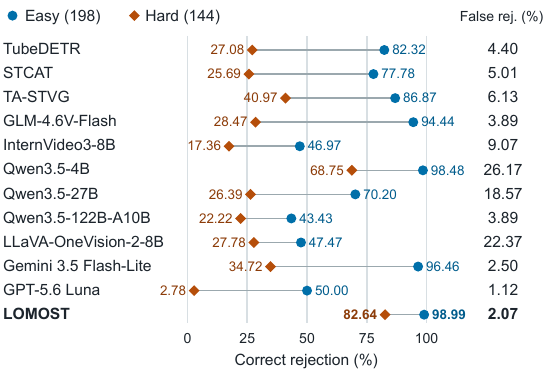}
\caption{Zero-target recognition on TEST1500. Circles show correct rejections for cross-video mismatches (easy), and diamonds show correct rejections for same-video counterfactuals (hard). The right column gives false rejection rates on 1,158 target-present queries.}
\label{fig:zero-split}
\end{minipage}
\end{figure*}
\paragraph{A1. Similar instance disambiguation.}
The query asks the model to localize the man in a black jacket and the man in a striped tracksuit walking side by side down a hallway. The intervals predicted by STCAT and TA-STVG have no overlap with the ground-truth interval, missing the target period entirely. Although TubeDETR roughly localizes the interval (36.97\% tIoU), its predicted instance count is inconsistent. The ground truth contains two boxes, but TubeDETR predicts three boxes at 2:16.50 and merges the two men into a single large box at 2:18.00. LOMOST predicts a more accurate interval than the baselines (60.00\% tIoU), consistently separates the two target men throughout the interval, and excludes the women mentioned as contextual references (37.63\% vIoU).

\paragraph{B1. Short target interval and shot transitions.}
All three baselines miss a box at 5:11.51. LOMOST achieves the highest tIoU (92.67\%) and consistently maintains two separate, accurate bounding boxes in every frame within the interval, achieving \textbf{82.82\% vIoU}.

\paragraph{C1. Spatial correction in motion.}
This example illustrates the roles of the Spatial Localization Calibrator (SQCM + FPCM) and Box Refinement. STCAT predicts an interval that entirely misses the ground-truth interval. TubeDETR correctly localizes the knife at 2:08.00, but at 2:07.50 and 2:08.50 its boxes remain in similar locations and cover the hand while missing the moving knife. TA-STVG produces several mislocalized boxes and fails to tightly enclose the knife at 2:08.00. For LOMOST, 2:08.00 is an anchor frame predicted by the global stage (marked \texttt{* anchor}). By 2:08.50, the knife slicing the onion has moved visibly downward. Directly copying the anchor box (Box-copy) yields an IoU of only 0.23 on this frame. After local calibration updates the target center and scale, the boxes at 2:07.50 and 2:08.50 move to the correct knife locations in their respective frames. The \textbf{refined box} at 2:08.50 reaches an IoU of 0.67, with a vIoU of 49.96\% over the full interval.

\paragraph{D1. Temporal localization failure.}
The interaction in which two people run quickly down a hospital hallway occupies only 1.31\% of the nine-minute video. We hypothesize that LOMOST's sparse global sampling at 0.5~FPS may have caused the VLM to miss this brief sequence and select an earlier, visually similar segment that does not match the query semantics.

As shown in Figure~\ref{fig:supp-spatial-calibration}, we use the example of a knife slicing an onion to show how the Spatial Localization Calibrator and the Box Refinement Head refine the coarse boxes predicted by the VLM. The top row shows the VLM's coarse boxes on three anchor frames. For the current frame at 2:08.50, we copy the coarse box from anchor frame 2, 0.50 seconds earlier, onto the current frame as the starting point for refinement. The bottom row shows the SQCM score \(\boldsymbol{S}_{\mathrm{context}}\), the FPCM score \(\boldsymbol{S}_{\mathrm{source}}\), and the fused score \(\boldsymbol{L}\) on the same candidate grid in the current frame. SQCM and FPCM produce different spatial responses on the candidate grid in the current frame. The peak of the fused score lies below and to the right of the copied box center, within the knife's ground-truth box. The red arrow in the rightmost panel marks the displacement from the center of the copied box to the center of the refined box. After refinement, the frame-level IoU with the same ground-truth box increases from 0.23 to 0.67.

\subsection{Spatial Refinement by Distance to the Anchor Frame}
\label{supp:anchor-distance}
We group the 2~FPS scoring frames in the intersection of the predicted and ground-truth intervals by their absolute time difference from the anchor used for prediction. Table~\ref{tab:supp-anchor-distance} includes all 19,495 frames with an identifiable anchor across 956 queries.

Spatial refinement primarily benefits non-anchor frames. With equal weight given to each query within a bin, FULL outperforms Box Copy in every non-anchor distance bin. The mean spatial IoU gain is 2.30 percentage points for non-anchor frames within 1~s of the anchor. The gain increases with distance up to a peak of 6.45 points in the 4--8~s bin and remains positive at 4.88 points beyond 8~s.

\begin{table}[!htbp]
    \centering
    \caption{Spatial IoU by distance to the anchor frame on LV-MOSTG-9K. We group scoring frames by their temporal distance from the anchor frame used for prediction. In each group, we compute the mean spatial IoU for each query and then average these means equally across queries. A query may contribute to multiple bins. Gain is FULL minus Box Copy in percentage points, computed before rounding.}
    \label{tab:supp-anchor-distance}
    \small
    \setlength{\tabcolsep}{3pt}
    \begin{tabular*}{\columnwidth}{@{\extracolsep{\fill}}lrrrrr@{}}
        \toprule
        & & & \multicolumn{2}{c}{Spatial IoU (\%)} & \\
        \cmidrule(lr){4-5}
        Distance (s) & Queries & Frames & Box Copy & FULL & Gain \\
        \midrule
        $0$     & 936 & 2,159 & 74.79 & 73.33 & $-1.45$ \\
        $(0,1]$ & 945 & 5,742 & 57.67 & 59.98 & $+2.30$ \\
        $(1,2]$ & 798 & 4,656 & 47.96 & 52.27 & $+4.31$ \\
        $(2,4]$ & 442 & 4,063 & 42.99 & 47.81 & $+4.81$ \\
        $(4,8]$ & 149 & 2,024 & 39.15 & 45.59 & $+6.45$ \\
        $>8$    &  29 &   851 & 39.94 & 44.82 & $+4.88$ \\
        \bottomrule
    \end{tabular*}
\end{table}

\subsection{Stability of the Spatial Localization Ablations}
\label{supp:spatial-stability}
\paragraph{Fine-tuned 4B VLM baseline.}
In the ablation study (Table~5 of the main paper), SFT + Box Copy uses the same fine-tuned Qwen3-VL-4B checkpoint as full LOMOST and copies boxes from the nearest anchor frame. For LV-MOSTG-9K, we also report results for \textbf{SFT + Linear} in Table~\ref{tab:supp-sft-linear}, using the same predicted intervals and anchor boxes with the linear interpolation protocol used for the VLM baselines in Table~2 of the main paper.

\begin{table}[!htbp]
    \centering
    \caption{Spatial grounding on LV-MOSTG-9K TEST1500 (\%). All three configurations share the same temporal predictions and zero-target decisions, with 49.34\% m-tIoU and 92.51\% zero-target F1.}
    \label{tab:supp-sft-linear}
    \footnotesize
    \setlength{\tabcolsep}{2pt}
    \begin{tabular*}{\columnwidth}{@{\extracolsep{\fill}}lrrrr@{}}
        \toprule
        Configuration & m-vIoU & \shortstack{vIoU-Int.\\All} & vIoU@0.3 & vIoU@0.5 \\
        \midrule
        SFT + Box Copy & 28.67 & 45.89 & 46.20 & 20.29 \\
        SFT + Linear   & 28.98 & 46.16 & 46.80 & 20.81 \\
        Full LOMOST    & \textbf{30.27} & \textbf{48.16} & \textbf{49.91} & \textbf{22.11} \\
        \bottomrule
    \end{tabular*}
\end{table}

Given the same predicted time intervals and anchor boxes, the second stage refines only spatial positions. Compared with directly copying the anchor boxes, the calibrator improves overall m-vIoU from 28.67\% to 30.27\% (\textbf{paired 95\% CI [1.27, 1.94]}). Since spatial refinement cannot correct temporal localization errors, we further examine localization quality where the predicted and ground-truth intervals overlap. The vIoU-Int.\ All score increases from 45.89\% to 48.16\% (\textbf{+2.27 percentage points}). When frames are grouped by their temporal distance from the anchor frame used for prediction, the calibrator outperforms Box Copy in every non-anchor group. The largest gain occurs in the group \textbf{4--8 seconds} from the anchor frame, where mean spatial IoU increases by \textbf{6.45 percentage points} across 149 queries and 2,024 frames.

FPCM-only and FULL share the frozen VLM, predicted intervals, anchor boxes, and training data. We train their localization modules separately for 1,242 steps and evaluate each configuration using its final checkpoint.

\paragraph{Paired bootstrap.}
With the weights from Table~5 of the main paper fixed, we perform a paired bootstrap over the 433 source videos represented by the 1,158 queries with matching targets. In each replicate, we sample source videos with replacement, use the same sample for all configurations, and compute differences in query-averaged m-vIoU. The 95\% percentile confidence intervals from 20,000 replicates lie above zero for all three comparisons (Table~\ref{tab:supp-spatial-bootstrap}).

\begin{table}[!htbp]
    \centering
    \caption{Paired differences in m-vIoU and their 95\% confidence intervals (CI), in percentage points.}
    \label{tab:supp-spatial-bootstrap}
    \small
    \begin{tabular*}{\columnwidth}{@{\extracolsep{\fill}}lrr@{}}
        \toprule
        Comparison & Difference & 95\% CI \\
        \midrule
        FULL $-$ Box Copy      & $+1.60$ & $[1.27, 1.94]$ \\
        FPCM-only $-$ Box Copy & $+1.10$ & $[0.75, 1.46]$ \\
        FULL $-$ FPCM-only     & $+0.50$ & $[0.23, 0.77]$ \\
        \bottomrule
    \end{tabular*}
\end{table}

\paragraph{Training variability.}
To assess whether the gain from combining FPCM and SQCM is consistent across training seeds, we report results for FPCM-only and FULL across three seeds with the VLM and its outputs fixed. FULL outperforms FPCM-only on all three seeds, with a mean paired m-vIoU gain of $0.63 \pm 0.12$ percentage points (mean $\pm$ sample standard deviation; Table~\ref{tab:supp-spatial-seeds}).

\begin{table}[!htbp]
    \centering
    \caption{Localization results across training seeds. FPCM-only and FULL scores are m-vIoU (\%). Gain is FULL minus FPCM-only in percentage points, computed before rounding. SD denotes the sample standard deviation across seeds.}
    \label{tab:supp-spatial-seeds}
    \footnotesize
    \setlength{\tabcolsep}{2pt}
    \begin{tabular*}{\columnwidth}{@{\extracolsep{\fill}}lrrr@{}}
        \toprule
        Training seed & FPCM-only & FULL & Gain \\
        \midrule
        20260917 (Tab.~5) & 29.771 & 30.271 & $+0.499$ \\
        20260918          & 29.782 & 30.427 & $+0.645$ \\
        20260919          & 29.862 & 30.608 & $+0.746$ \\
        \midrule
        Mean $\pm$ SD & $29.805 \pm 0.050$ & $30.435 \pm 0.169$ & $+0.630 \pm 0.124$ \\
        \bottomrule
    \end{tabular*}
\end{table}

\subsection{Limitations and Future Work}
\label{supp:limitations}
The current version of LV-MOSTG-9K supports only one continuous target time interval per expression with matching targets. When the same expression has matching targets in multiple disjoint intervals of a video, the benchmark cannot yet assess whether a model can identify all such intervals. LOMOST sparsely samples the full video at 0.5~FPS and may miss brief visual cues needed to identify the target interval. During local grounding, the Spatial Localization Calibrator searches for target centers only near the anchor boxes, and the current frame inherits the number of boxes predicted by the VLM at the nearest anchor frame. When a target moves outside the search region or the number of visible targets changes between anchor frames, adjusting only the positions and sizes of existing boxes makes it difficult to localize all visible targets accurately. Future work could explore datasets for multi-interval spatio-temporal grounding and methods that update the set of visible targets frame by frame.

\end{document}